\pdfoutput=1

\documentclass[runningheads]{llncs}

\usepackage[mobile]{eccv}

\usepackage{eccvabbrv}

\usepackage{graphicx}
\usepackage{booktabs}

\usepackage[accsupp]{axessibility}  

\usepackage[table]{colortbl}
\usepackage{verbatim}
\usepackage{setspace}

\usepackage{hyperref}

\usepackage{orcidlink}

\usepackage{dsfont}
\usepackage{etoolbox}
\usepackage{color}

\newif\ifshowedits

\newcommand{\addeditor}[3]{%
  \definecolor{#1color}{rgb}{#3}
  \expandafter\newcommand\csname #1\endcsname[1]{%
  \ifshowedits
    {\color{#1color} ##1}%
  \else
    {##1}%
  \fi
  }%
  \expandafter\newcommand\csname #1rmk\endcsname[1]{%
  \ifshowedits
    {\color{#1color} {\bf [#2: ##1]}}
  \fi
  }%
  \expandafter\newcommand\csname #1rpl\endcsname[2]{%
  \ifshowedits
    {\color{#1color} ##1 \sout{##2}}
  \else
    {##1}
  \fi
  }%
}

\newcommand{\createtextvar}[1]{
  \expandafter\newcommand\csname #1\endcsname{%
  {\text{#1}}
}%
}
\newcommand{\textvars}[1]{\forcsvlist{\createtextvar}{#1}}

\newcommand{\moretextwithfigures}{
\renewcommand{\topfraction}{1}
\renewcommand{\dbltopfraction}{1}
\renewcommand{\bottomfraction}{1}
\renewcommand{\textfraction}{.0}
\renewcommand{\floatpagefraction}{1}
\renewcommand{\dblfloatpagefraction}{1}
}

\newcommand{\mycomment}[1]{}

\newcommand{\calG}{{\cal G}}

\newcommand{\calM}{{\cal M}}

\newcommand{\IN}{{\mathds{N}}}

\newcommand{\IR}{{\mathds{R}}}

\newcommand{\vcomment}[1]{}

\textvars{out,half}

\newcommand{\In}{\text{in}}
\newcommand{\Mid}{\text{mid}}

\newcommand{\innershift}{\delta^\In}
\newcommand{\outershift}{\delta^\out}
\newcommand{\propinnershift}{\epsilon^\In}
\newcommand{\propoutershift}{\epsilon^\out}

\usepackage{adjustbox}

\addeditor{vincent}{VL}{0.0, 0.5, 0.0}
\addeditor{antoine}{AG}{0.0, 0.0, 0.8}
\showeditstrue
\showeditsfalse
\moretextwithfigures

\begin{document}


\title{Gaussian Frosting: Editable Complex Radiance Fields with Real-Time Rendering}

\titlerunning{Gaussian Frosting}

\author{
\vspace{-0.9cm} \> \\
Antoine Guédon\orcidlink{0009-0001-3107-4454} \and
Vincent Lepetit\orcidlink{0000-0001-9985-4433}
}

\authorrunning{A.~Guédon and V.~Lepetit}

\institute{\vspace{-0.5cm} \> \\
\email{firstname.lastname@enpc.fr}\\
LIGM, Ecole des Ponts, Univ Gustave Eiffel, CNRS, France\\
{\tt\small \url{https://anttwo.github.io/frosting/}}
}

\maketitle

\begin{abstract}
\vspace{-1.5cm}
\begin{figure}[ht]
  \centering
  \begin{subfigure}{0.31\linewidth}
  \includegraphics[width=\linewidth]{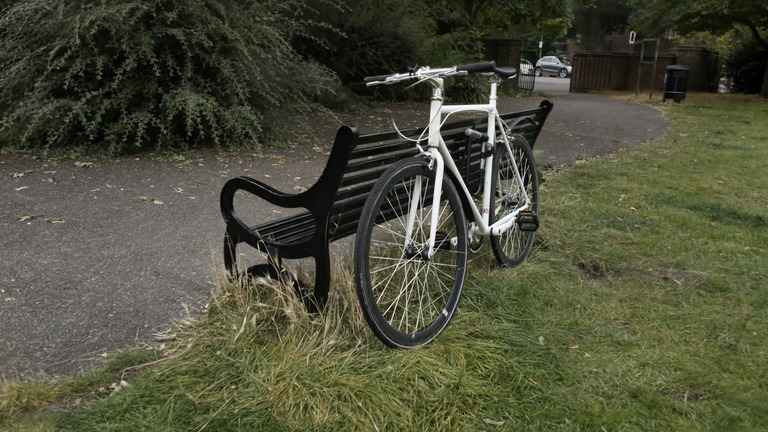}
  \end{subfigure}
  \hfill
  \begin{subfigure}{0.31\linewidth}
  \includegraphics[width=\linewidth]{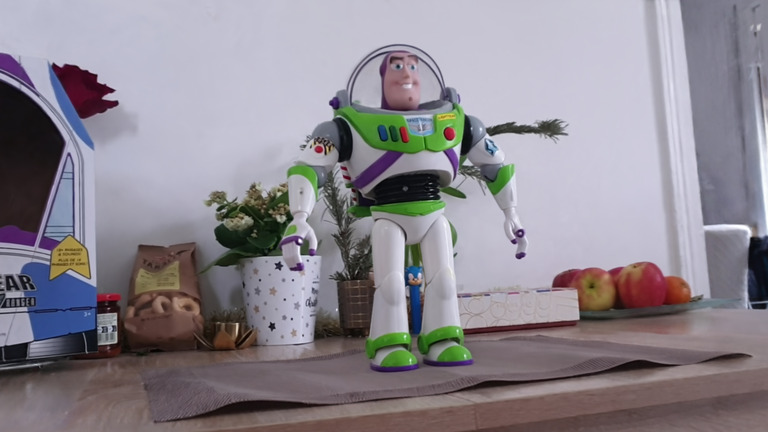}
  \end{subfigure}
  \hfill
  \begin{subfigure}{0.31\linewidth}
  \includegraphics[width=\linewidth]{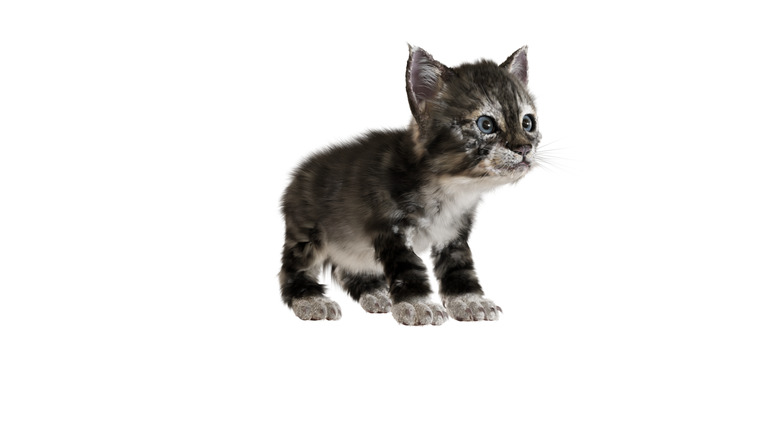}
  \end{subfigure}\\
  {\small (a)~Rendering three different scenes with Frosting: \textit{Bicycle}, \textit{Buzz}, and \textit{Kitten}.}\\
  \begin{subfigure}{0.49\linewidth}
  \includegraphics[width=\linewidth]{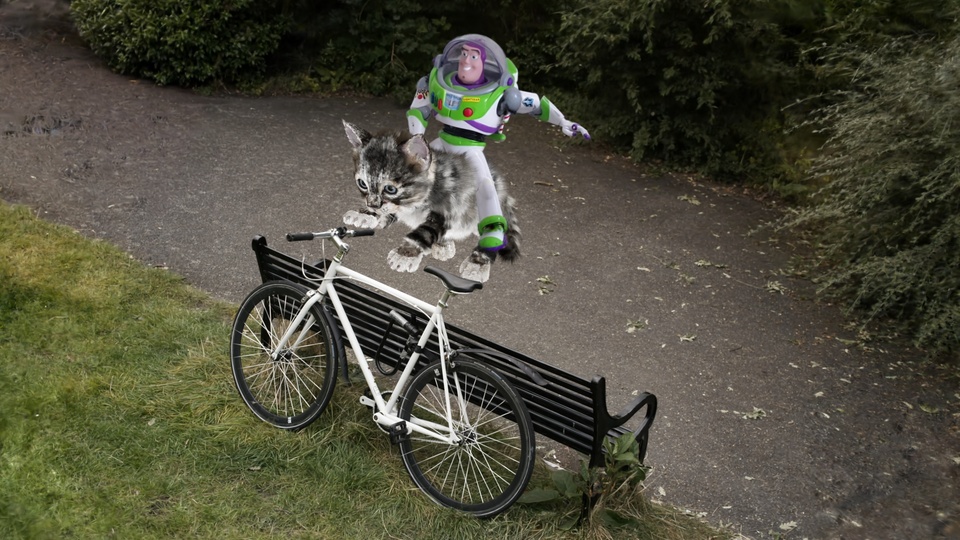}
  \includegraphics[width=\linewidth]{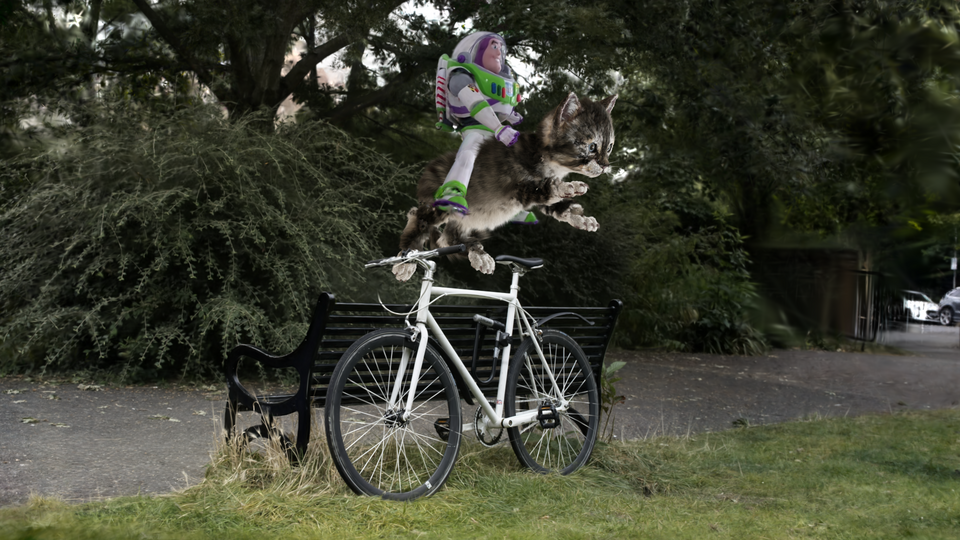}
  \end{subfigure}
  \hfill
  \begin{subfigure}{0.49\linewidth}
  \includegraphics[width=\linewidth]{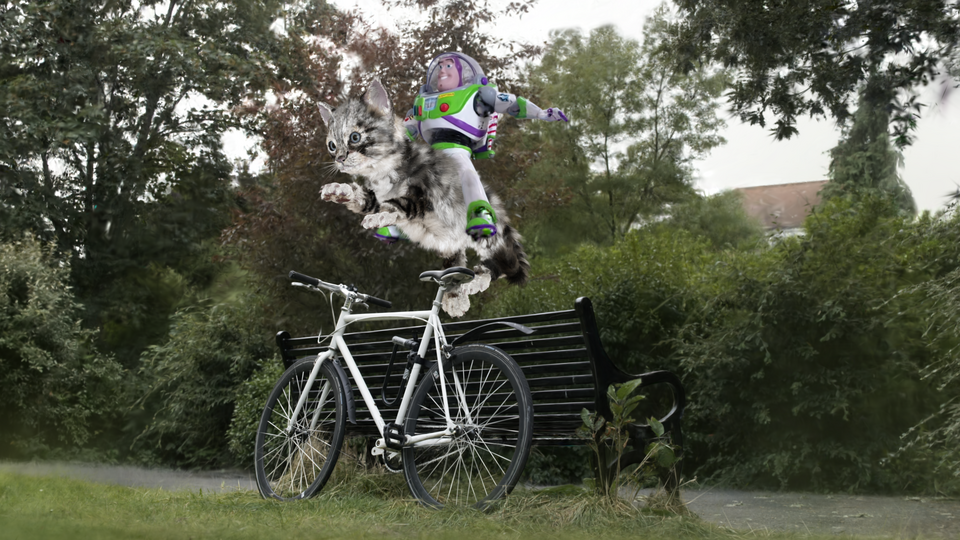}
  \includegraphics[width=\linewidth]{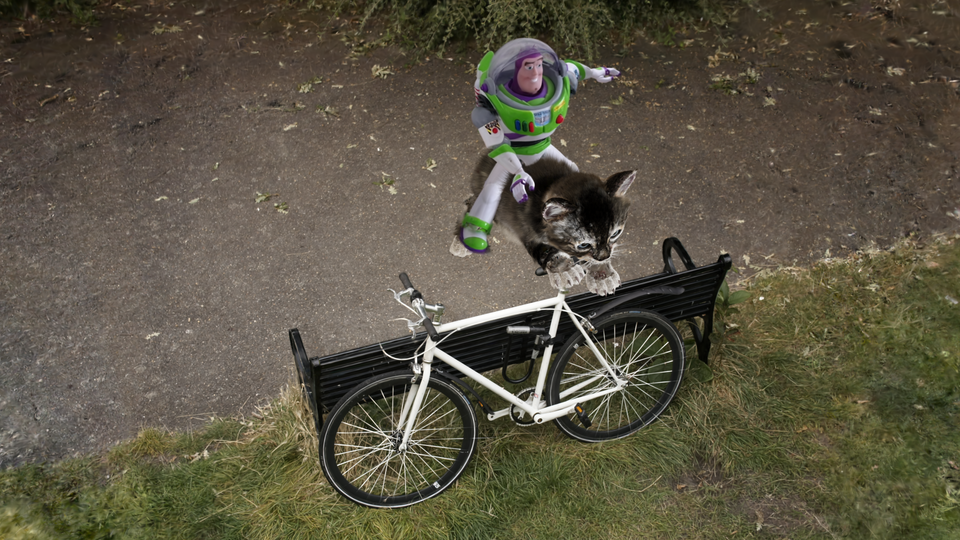}
  \end{subfigure}
  {\small (b)~Composition: \textit{Buzz is riding a giant kitten jumping over a bench.}}\\
  \begin{subfigure}{0.49\linewidth}
 \includegraphics[width=0.49\linewidth]{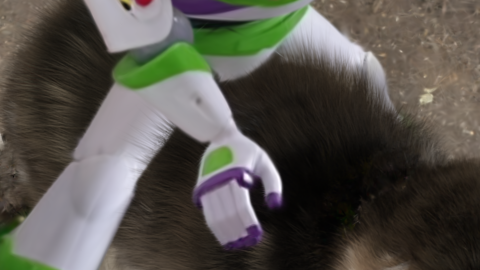}
 \includegraphics[width=0.49\linewidth]{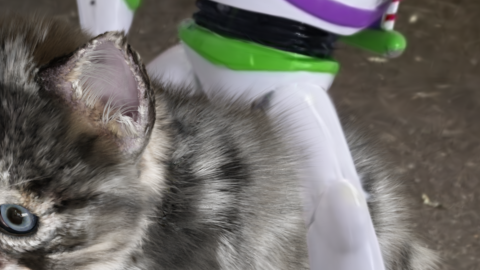}
 \caption*{(c) Fuzzy details rendered with Frosting - occlusions are correctly rendered}
  \end{subfigure}
  \hfill
  \begin{subfigure}{0.49\linewidth}
  \includegraphics[width=0.49\linewidth]{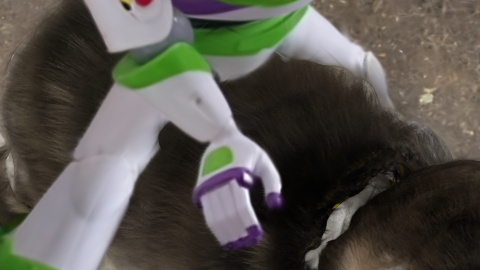}
 \includegraphics[width=0.49\linewidth]{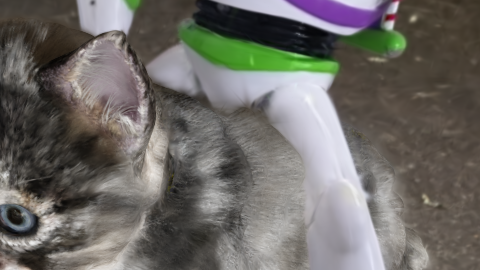}
  \caption*{(d) Rendering with SuGaR~\cite{guedon2023sugar} - the fur does not occlude the legs correctly}
  \end{subfigure}
  \caption{
  We propose to represent surfaces by a mesh covered with a ``Frosting'' layer of varying thickness and made of 3D Gaussians. This representation captures both complex volumetric effects created by fuzzy materials such as the cat's hair or grass as well as flat surfaces. 
  Built from RGB images only, it can be rendered in real-time and animated using traditional animation tools.
  In the example above, we were able to animate both Buzz and the kitten, changing their original pose (a) while preserving high-quality rendering (b): Contrary to SuGaR, very fine and fuzzy details such as the kitten's hair can be seen covering Buzz's legs in a realistic way (c).
  }
  \label{fig:sugar-comparison}
\end{figure}
\begin{figure}[h]
  \centering

   \begin{subfigure}{0.49\linewidth}
  \includegraphics[width=0.99\linewidth]{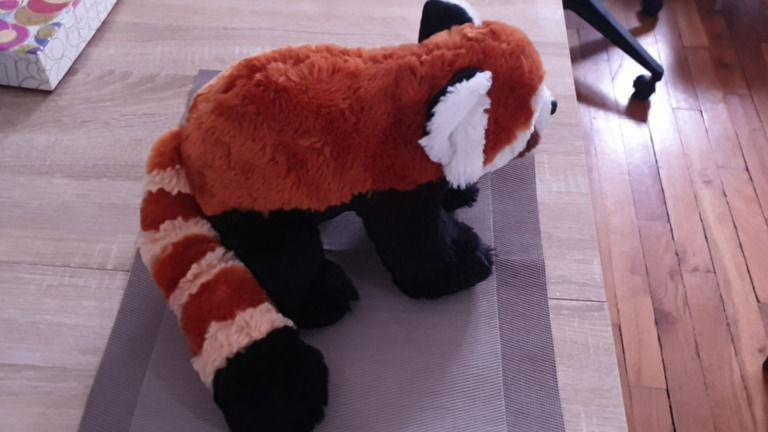}
  \caption{Rendering}
  \end{subfigure}
  \hfill
  \begin{subfigure}{0.49\linewidth}
  \includegraphics[width=0.99\linewidth]{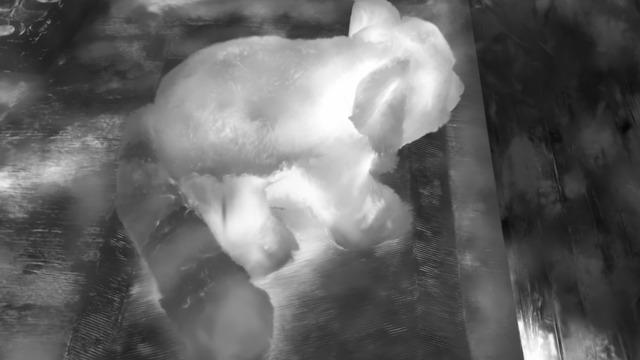}
  \caption{Thickness of the Frosting layer}
  \end{subfigure}
  
  \caption{
  \textbf{Visualization of the thickness of the Frosting layer on an example.} A thicker layer is shown with a brighter value. Our method automatically builds a thick Frosting layer for fuzzy areas  such as the fur of the red panda plush, and a thin Frosting layer for flat surfaces such as the table or the floor. Adapting the thickness of the Frosting layer allows for allocating more Gaussians in areas where more volumetric rendering is necessary near the surface, resulting in an efficient distribution of Gaussians in the scene. As we demonstrate in the paper, using an adaptive thickness results in higher performance than using a predefined constant thickness, and reduces artifacts when animating the mesh.}
  \label{fig:teaser}
\end{figure}
We propose Gaussian Frosting, a novel mesh-based representation for high-quality rendering and editing of complex 3D effects in real-time. Our approach builds on the recent 3D Gaussian Splatting framework, which optimizes a set of 3D Gaussians to approximate a radiance field from images. We propose first extracting a base mesh from Gaussians during optimization, then building and refining an adaptive layer of Gaussians with a variable thickness around the mesh to better capture the fine details and volumetric effects near the surface, such as hair or grass. We call this layer Gaussian Frosting, as it resembles a coating of frosting on a cake. The fuzzier the material, the thicker the frosting. We also introduce a parameterization of the Gaussians to enforce them to stay inside the frosting layer and automatically adjust their parameters when deforming, rescaling, editing or animating the mesh. Our representation allows for efficient rendering using Gaussian splatting, as well as editing and animation by modifying the base mesh. We demonstrate the effectiveness of our method on various synthetic and real scenes, and show that it outperforms existing surface-based approaches. We will release our code and a web-based viewer as additional contributions.
  
\end{abstract}

\vspace{1cm}
\section{Introduction}

3D Gaussian Splatting~(3DGS)~\cite{kerbl3Dgaussians} has recently conquered the field of 3D reconstruction and image-based rendering. By representing a scene with a large set of tiny Gaussians, 3DGS allows for fast reconstruction and rendering, while nicely capturing fine details and complex light effects. Compared to earlier neural rendering methods such as NeRFs~\cite{mildenhall2020nerf}, 3DGS is 
much more efficient for both the reconstruction and rendering stages by a large margin.

However, like NeRFs, Vanilla 3DGS does not allow easy edition of the reconstructed scene. The Gaussians are unstructured, disconnected from each other, and it is not clear how a designer can manipulate them, for example to animate the scene. Very recently, SuGaR~\cite{guedon2023sugar} showed how to extract a mesh from the output of 3DGS. Then, by constraining Gaussians to stay on the mesh, it is possible to edit the scene with traditional Computer Graphics tools for manipulating meshes. But by flattening the Gaussians onto the mesh, SuGaR loses the rendering quality possible with 3DGS for fuzzy materials and volumetric effects.

In this paper, we introduce Gaussian Frosting--or Frosting, for short--a hybrid representation of 3D scenes that is editable as a mesh while providing a rendering quality at least equal, sometimes superior to 3DGS. The key idea of Frosting is to augment a mesh with a layer containing Gaussians. The thickness of the Frosting layer adapts locally to the material of the scene: The layer should be thin for flat surfaces to avoid undesirable volumetric effects, and thicker around fuzzy materials for realistic rendering. As shown in Figure~\ref{fig:teaser}, using the Frosting representation, we can not only retrieve a highly accurate editable mesh but also render complex volumetric effects in real-time.

Frosting is reminiscent of the Adaptive Shells representation~\cite{wang-siggraphasia2023-adaptive-shells}, which relies on two explicit triangle meshes extracted from a Signed Distance Function to control volumetric effects. Still, Frosting allies a rendering quality superior to the quality of Adaptive Shells to the speed efficiency of 3DGS for reconstruction and rendering, while being easily editable as it depends on a single mesh.

While the Frosting representation is simple, it is challenging to define the local thickness of its layer. To extract the mesh, we essentially rely on SuGaR, which we improved with a technique~(described in the supplementary material) to automatically tune a critical hyperparameter. To estimate the Frosting thickness, we introduce a method to define an inner and outer bound for the Frosting layer at each vertex of this mesh based on the Gaussians initially retrieved by 3DGS around the mesh.  Finally, we populate the Frosting layer with randomly sampled Gaussians and optimize these Gaussians constrained to stay within the layer. We also propose a simple method to automatically adjust in real-time the parameters of the Gaussians when animating the mesh.

In summary, we propose a simple yet powerful surface representation that captures complex volumetric effects, can be edited with traditional tools, and can be rendered in real-time. We also propose a method to build this representation from images, based on recent developments on Gaussian Splatting. We will release our code including a viewer usable in browser as an additional contribution. 
\begin{figure}[ht]
  \centering
  \begin{subfigure}{0.14\linewidth}
 \includegraphics[width=\linewidth]{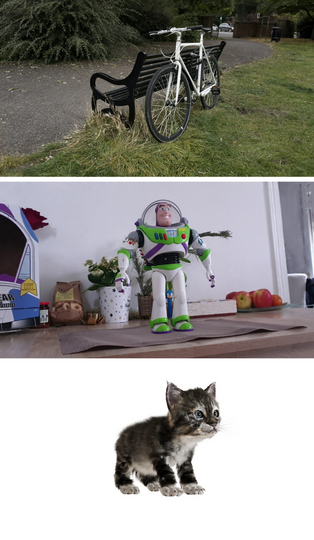}
 \includegraphics[width=\linewidth]{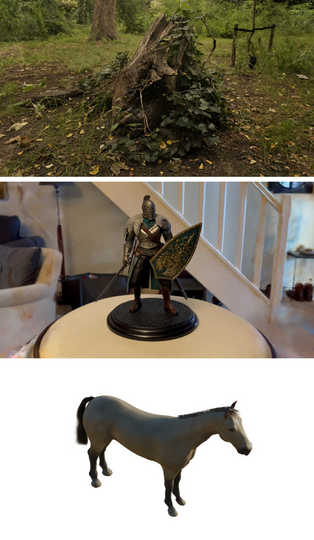}
 \caption{Scenes}
  \end{subfigure}
  \hfill
  \begin{subfigure}{0.42\linewidth}
 \includegraphics[width=\linewidth]{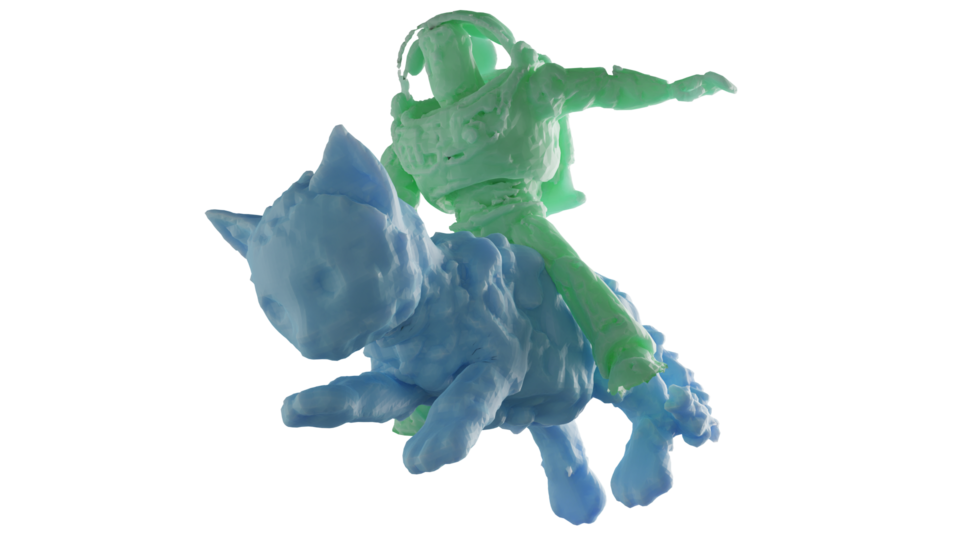}
 \includegraphics[width=\linewidth]{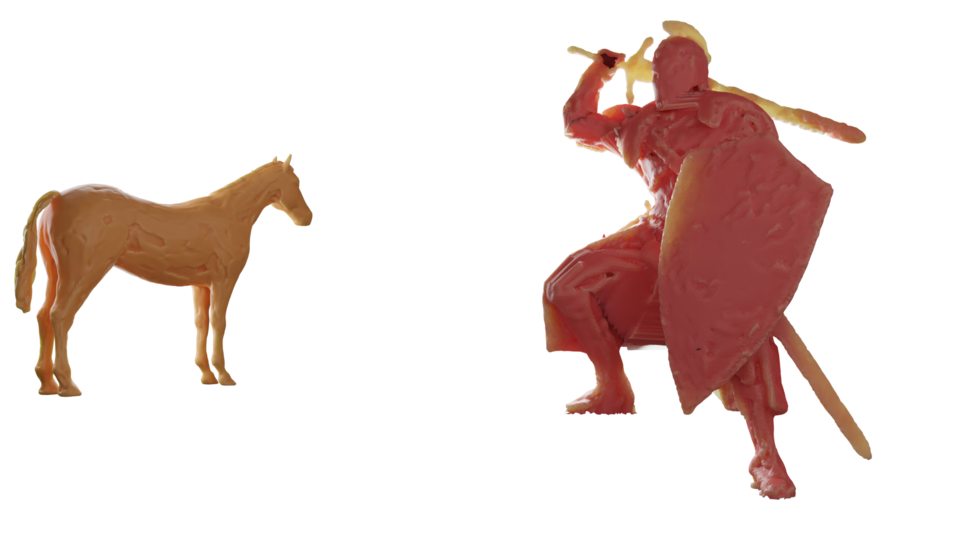}
  \caption{Posing foreground meshes}
  \end{subfigure}
  \hfill
  \begin{subfigure}{0.42\linewidth}
 \includegraphics[width=\linewidth]{images/composition/buzz_riding_cat/0_0.png}
 \includegraphics[width=\linewidth]{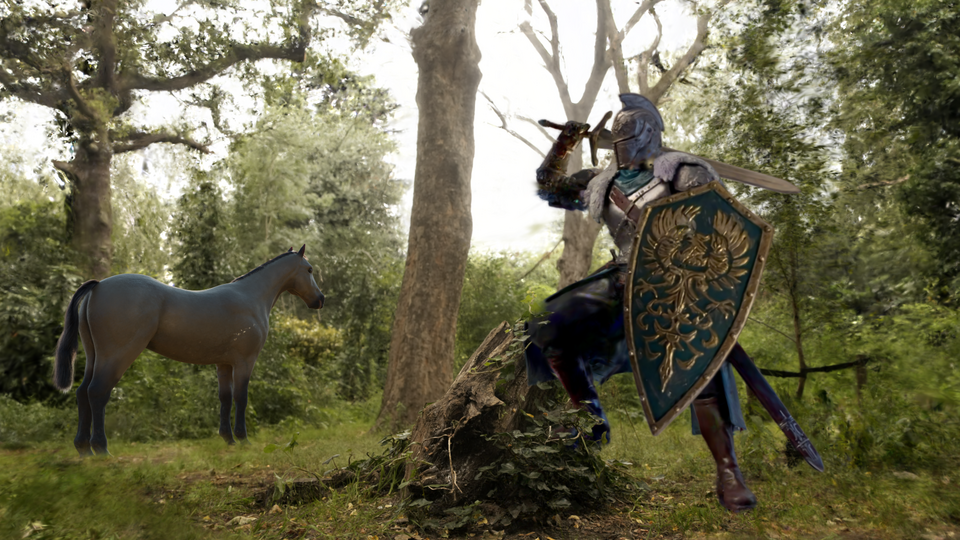}
  \caption{Rendering composition}
  \end{subfigure}
  \caption{
  \textbf{Scene composition.} Using mesh editing tools in Blender, we were able to combine various elements from multiple scenes (a) to build a whole new scene (c). We also changed the pose of the characters by using the rigging tool in Blender (b). Similarly to surface-based methods like SuGaR~\cite{guedon2023sugar}, Frosting can be used for editing and compositing scenes, but allows for better rendering of complex volumetric effects and fuzzy materials, such as hair or grass.
  }
  \label{fig:scene-composition}
\end{figure}

\section{Related Work}

The goal of image-based rendering (IBR) is to create a representation of a scene from a given set of images in order to generate new images of the scene. Different types of scene representations have been proposed, ranging from explicit and editable ones like triangle meshes or point clouds, to implicit or non-editable ones like voxel grids, multiplane images, or neural implicit functions. 

\noindent
\textbf{Volumetric IBR methods.} A recent breakthrough in IBR is Neural Radiance Fields (NeRF)~\cite{mildenhall2020nerf}, which uses a multilayer perceptron (MLP) to model a continuous volumetric function of density and color. NeRF can render novel views with high quality and view-dependent effects, by using volumetric ray tracing. However, NeRF is slow and memory hungry. 
Several works have tried to improve NeRF’s efficiency and training speed by using discretized volumetric representations like voxel grids and hash tables to store learnable features that act as inputs for a much smaller MLP~\cite{chen-eccv-2022-tensorf, karnewar2022relu, mueller2022instantngp, sun2022direct, yu_and_fridovichkeil2021plenoxels}, or to improve rendering performance by using hierarchical sampling strategies~\cite{barron2022mipnerf360,hedman2021snerg,reiser2021kilonerf,yu2021plenoctrees}. Other works have also proposed to modify NeRF's representation of radiance and include an explicit lighting model to increase the rendering quality for scenes with specular materials~\cite{verbin2022ref, boss2021nerd, kuang2022neroic, srinivasan2021nerv, zhang2021physg}. 
However, most volumetric methods rely on implicit representations that are not suited to editing compared to triangle meshes, for which most standard graphics hardware and software are tailored.

\noindent
\textbf{Surface-based IBR methods.} Triangle meshes have been a popular 3D representation for generating novel views of scenes~\cite{wood:2000:slf,buehler2001unstructured,hedman-2018-deepblending} after Structure-from-motion (SfM)~\cite{snavely-2006-structure-from-motion} and multi-view stereo (MVS)~\cite{goesele-2007-multiviewstereo} have enabled 3D reconstruction of surfaces. 
Deep-learning-based mesh representations~\cite{riegler2020free,riegler2021stable} have also been used for improving view synthesis using explicit surface meshes; However, even though mesh-based methods allow for very efficient rendering, they have trouble capturing complex and very fine geometry as well as fuzzy materials.

\noindent
\textbf{Hybrid IBR methods.} Some methods use a hybrid volumetric representation to recover surface meshes that are suitable for downstream graphics applications while efficiently modeling view-dependent appearance. 
Specifically, some works optimize a Neural Radiance Field in which the density is replaced by an implicit signed distance function (SDF), which provides a stronger regularization on the underlying geometry~\cite{oechsle2021unisurf,yariv2021volsdf,wang2021neus,li-cvpr2023-neuralangelo,darmon-2022-warp,bao-2022-neumesh}. However, most of these methods are not aimed at real-time rendering.
To mitigate this issue, other approaches greatly accelerate rendering by ``baking'' the rendering computation into the extracted mesh after optimization with a dedicated view-dependent appearance model~\cite{chen2022mobilenerf,yariv-2023-bakedsdf,reiser2024binaryopacitygrid}. 
Even though these surface-based methods encode the surface using a volumetric function represented by an MLP during optimization, they struggle in capturing fine details or fuzzy materials compared to volumetric methods.

Adaptive Shells~\cite{wang-siggraphasia2023-adaptive-shells} is a recent method that achieves a significant improvement in rendering quality by using a true hybrid surface-volumetric approach that restricts the volumetric rendering of NeRFs to a thin layer around the object. This layer is bounded by two explicit meshes, which are extracted after optimizing an SDF-based radiance field. The layer’s variable thickness also improves the rendering quality compared to a single flat mesh. This method combines the high-quality rendering of a full volumetric approach with the editability of a surface-based approach by manipulating the two meshes that define the layer. However, Adaptive Shells depends on a neural SDF~\cite{wang2021neus}, which has some limitations in its ability to reconstruct precise surfaces, and requires more than 8 hours to optimize a single synthetic scene, which is much longer than the recent Gaussian Splatting methods. 

\noindent
\textbf{Gaussian Splatting.} Gaussian Splatting~\cite{kerbl3Dgaussians} is a new volumetric representation inspired by point cloud-based radiance fields~\cite{kopanas2021point, ruckert2021adop} which is very fast to optimize and allows for real-time rendering with very good quality. One of its greatest strengths is its explicit 3D representation, which enables editing tasks as each Gaussian exists individually and can be easily adjusted in real-time. 
Some appearance editing and segmentation methods have been proposed~\cite{chen2023gaussianeditor,huang2023pointn,chungmin-2024-garfield,ye-2023-gaussian_grouping}, but the lack of structure in the point cloud makes it almost impossible for a 3D artist or an animator to easily modify, sculpt or animate the raw representation. The triangle mesh remains the standard 3D structure for these applications.
A recent work, SuGaR~\cite{guedon2023sugar}, extends this framework by aligning the Gaussians with the surface and extracting a mesh from them. Gaussians are finally flattened and pinned on the surface of the mesh, which provides a hybrid representation combining the editability of a mesh with the high-quality rendering of Gaussian Splatting. However, SuGaR remains a surface-based representation with limited capacity in reconstructing and rendering fuzzy materials and volumetric effects, resulting in a decrease in performance compared to vanilla Gaussian Splatting.
\section{3D Gaussian Splatting and Surface Reconstruction}

Our method relies on the original 3D Gaussian Splatting~(3DGS) method~\cite{kerbl3Dgaussians} for initialization and on SuGaR~\cite{guedon2023sugar} to align Gaussians with the surface of the scene and facilitate the extraction of a mesh. We briefly describe 3DGS and SuGaR in this section before describing our method in the next section.

\subsection{3D Gaussian Splatting} 

3DGS represents the scene as a large set of Gaussians. Each Gaussian $g$ is equipped with a mean $\mu_g\in \IR^3$ and a positive-definite covariance matrix $\Sigma_g\in \IR^{3\times 3}$. The covariance matrix is parameterized by a scaling vector $s_g\in\IR^3$ and a quaternion $q_g\in\IR^4$ encoding the rotation of the Gaussian. 

In addition, each Gaussian has a view-dependent radiance represented by an opacity $\alpha_g\in [0,1]$ and a set of spherical harmonics coordinates defining the colors emitted for all directions. To render an image from a given viewpoint, a rasterizer ``splats'' the 3D Gaussians into 2D Gaussians parallel to the image plane and blends the splats depending on their opacity and depth. This rendering is extremely fast, which is one of the advantages of 3DGS over volumetric rendering as in NeRFs for example~\cite{mildenhall2020nerf, mueller2022instantngp, barron2022mipnerf360}.

Gaussian Splatting can be seen as an approximation of the traditional volumetric rendering of radiance fields with the following density function $d$, computed as the sum of the Gaussian values weighted by their alpha-blending coefficients at any 3D point $p\in \IR^3$:
 \begin{equation}
    d(p) = \sum_{g} \alpha_g \exp\left(-\frac{1}{2}(p - \mu_g)^T \Sigma^{-1}_g (p - \mu_g)\right) \> .
    \label{eq:gaussian_splatting_density}
\end{equation}

We initialize our Gaussian Frosting method using a vanilla 3DGS optimization: Gaussians are initialized using the point cloud produced by an SfM~\cite{snavely-2006-structure-from-motion} algorithm like COLMAP~\cite{schoenberger2016mvs,schoenberger2016sfm}, required to compute camera poses. The Gaussians' parameters (3D means, scaling vectors, quaternions, opacities, and spherical harmonics coordinates) are then optimized to make the renderings match the ground truth images of the scene, using a rendering loss that only consists in a combination of a pixel-wise L1 distance and a more structural D-SSIM term.

\subsection{SuGaR Mesh Extraction} 

Vanilla 3DGS does not have regularization explicitly encouraging Gaussians to align with the true surface of the scene. Our Gaussian Frosting representation relies on a mesh that approximates this surface, in order to be editable by traditional tools. To obtain this mesh, we rely on the method proposed in SuGaR~\cite{guedon2023sugar}, which we improve by automatically selecting a critical hyperparameter.

SuGaR proposes a regularization term encouraging the alignment of the 3D Gaussians with the true surface of the scene during the optimization of Gaussian Splatting, as well as a mesh extraction method. After enforcing the regularization, the optimization provides Gaussians that are mostly aligned with the surface albeit not perfectly: We noticed that in practice, a large discrepancy between the regularized Gaussians and the extracted mesh indicates the presence of fuzzy materials or surfaces that require volumetric rendering. We thus exploit this discrepancy as a cue to evaluate where the Frosting should be thicker.

\begin{figure}[tb]
  \centering
  \includegraphics[width=\linewidth]{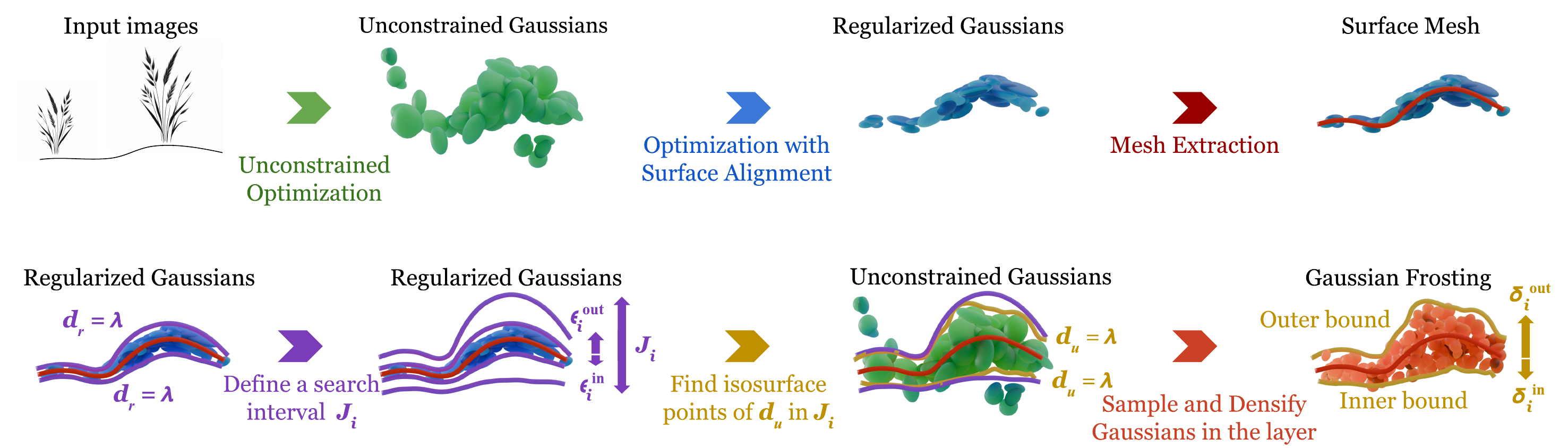}
  \caption{
  \textbf{Creating a Layer of Gaussian Frosting.} To build our proposed Frosting representation, we start by optimizing a Gaussian Splatting representation using a rendering loss without any additional constraint, to let Gaussians position themselves. We refer to these Gaussians as \emph{unconstrained}. We then regularize these Gaussians to enforce their alignement with the surface, and extract a mesh that will serve as a basis for the Frosting. Next, we use the misalignment of surface-aligned Gaussians to identify areas where more volumetric rendering is needed, and we build search intervals $J_i$ around the mesh's vertices $\vec{v_i}$. Finally, we use the density function of the unconstrained Gaussians to refine the intervals, resulting in a Frosting layer. We finally sample a novel, densified set of Gaussians inside the layer.
}
  \label{fig:frosting-pipeline}
\end{figure}

\section{Creating a Frosting Layer from Images}

In this section, we describe our Gaussian Frosting creation method: 
First, we extract an editable surface with optimal resolution using SuGaR. We then detail how we use this surface-based model to go back to a volumetric but editable representation built around the mesh. This representation adapts to the complexity of the scene and its need for more volumetric effects. Finally, we describe how we parameterize and refine this representation. An overview is provided Figure~\ref{fig:frosting-pipeline}.

\subsection{Forward Process: From Volume to Surface}
\label{subsec:forward-process}

We start by optimizing an unconstrained Gaussian Splatting representation for a short period of time to let Gaussians position themselves. We will refer to such Gaussians as \emph{unconstrained}. We save these Gaussians aside, and apply the regularization term from SuGaR to enforce the alignment of the Gaussians with the real surface. We will refer to these Gaussians as \emph{regularized}.

Once we obtain the regularized Gaussians, we extract a surface mesh from the Gaussian Splatting representation. This surface mesh serves as a basis for our representation. Like SuGaR~\cite{guedon2023sugar}, we then sample points on the visible level set of the Gaussian splatting density function, and apply Poisson reconstruction. 

\begin{figure}[tb]
  \centering
  \begin{subfigure}{0.19\linewidth}
  \includegraphics[width=\linewidth]{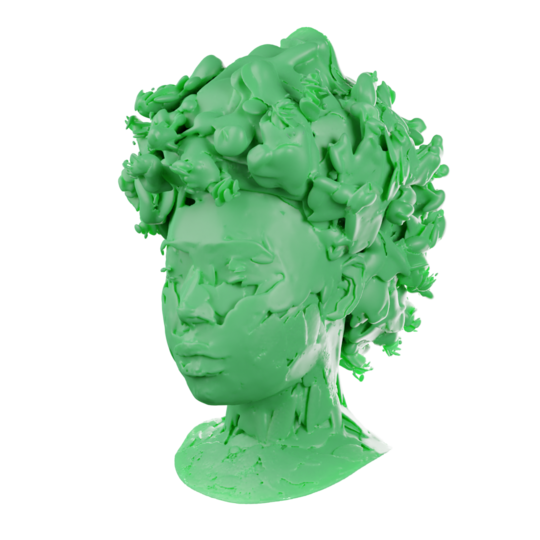}
  \end{subfigure}
  \hfill
  \begin{subfigure}{0.19\linewidth}
  \includegraphics[width=\linewidth]{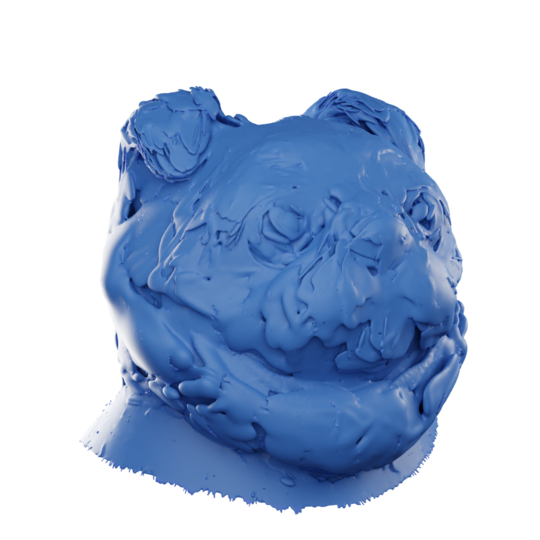}
  \end{subfigure}
  \hfill
  \begin{subfigure}{0.19\linewidth}
  \includegraphics[width=\linewidth]{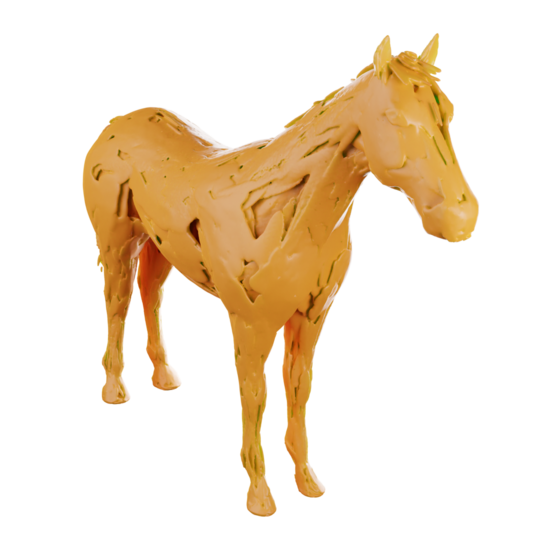}
  \end{subfigure}
  \hfill
  \begin{subfigure}{0.19\linewidth}
  \includegraphics[width=\linewidth]{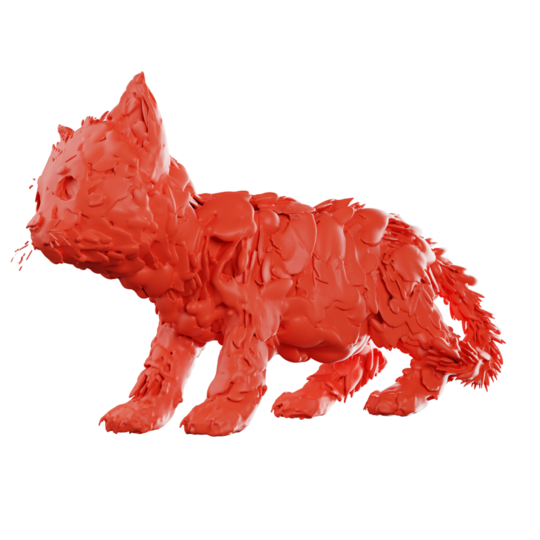}
  \end{subfigure}
  %
  %
  %
  \vspace{0.005\linewidth}\\
  {\small (a) Using the predefined, large parameter $D$ as in SuGaR~\cite{guedon2023sugar}} 
  \\
  \begin{subfigure}{0.19\linewidth}
  \includegraphics[width=\linewidth]{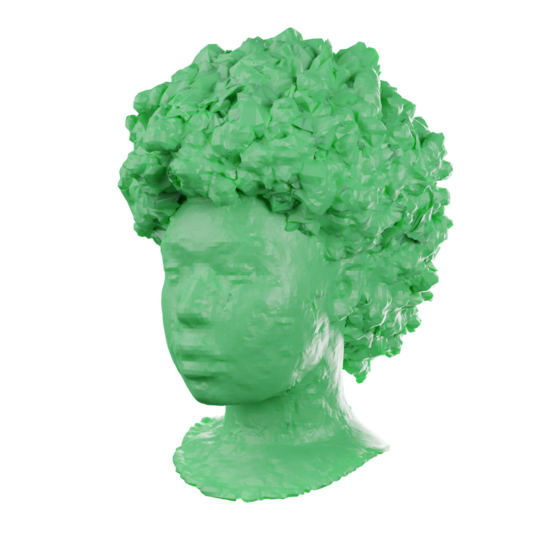}
  \end{subfigure}
  \hfill
  \begin{subfigure}{0.19\linewidth}
  \includegraphics[width=\linewidth]{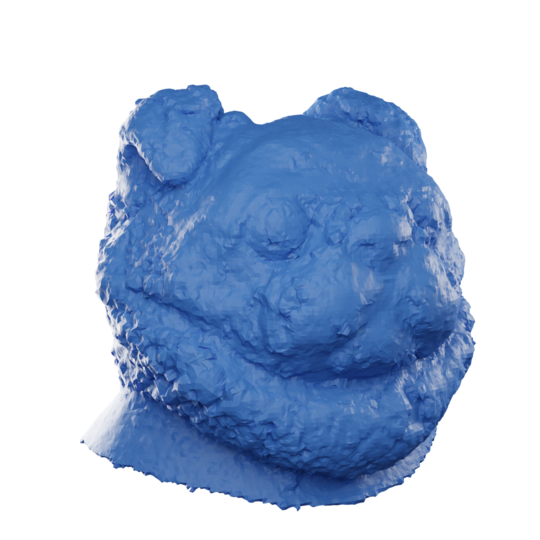}
  \end{subfigure}
  \hfill
  \begin{subfigure}{0.19\linewidth}
  \includegraphics[width=\linewidth]{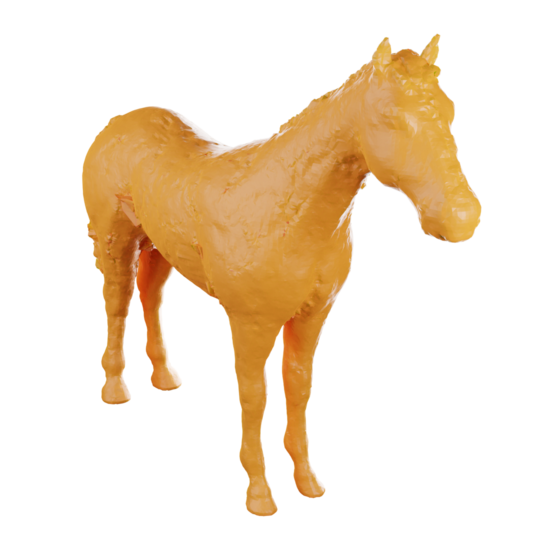}
  \end{subfigure}
  \hfill
  \begin{subfigure}{0.19\linewidth}
  \includegraphics[width=\linewidth]{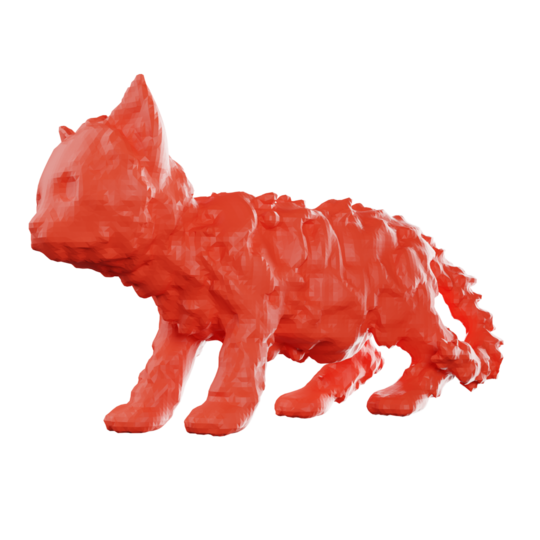}
  \end{subfigure}
  %
  %
  \\
  {\small (b) Using our automatically computed $D$ that adapts to the complexity of the 3DGS}
  \caption{
  \textbf{Comparison of meshes extracted by SuGaR from the Shelly dataset without and with our improvement that automatically tunes the octree depth $D$ in Poisson reconstruction depending on the complexity of the scene.}
  Our technique~(bottom) drastically reduces surface artifacts for many scenes, such as the holes and the ellipsoidal bumps on the surface when using the default values from~\cite{guedon2023sugar}~(top).
}

  \label{fig:mesh-comparison}
\end{figure}

In the supplementary material, we describe our technique to automatically estimate a good value for a critical hyperparameter used by Poisson reconstruction, namely the octree depth $D$. As we will show in the Experiments section, selecting the right value for $D$ when applying Poisson reconstruction can drastically improve both the quality of the mesh and the rendering performance of our model. Figure~\ref{fig:mesh-comparison} illustrates this point.

\subsection{Backward Process: From Surface to Volume}
\label{sec:shifts}

After extracting a base mesh, we build a Frosting layer with a variable thickness and containing Gaussians around this mesh. We want this layer to be thicker in areas where more volumetric rendering is necessary near the surface, such as fuzzy material like hair or grass for example. On the contrary, this layer should be very thin near the parts of the scene that corresponds to well-defined flat surfaces, such as wood or plastic for example.

\begin{figure}[t]
  \centering
  \includegraphics[width=\linewidth]{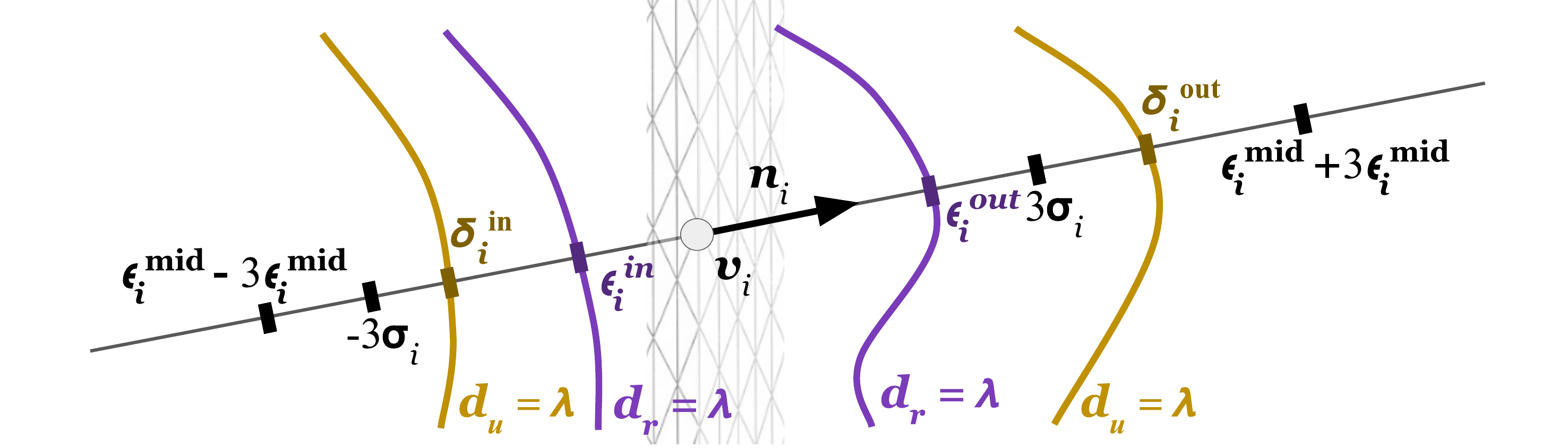}
  \caption{
  \textbf{How we define the inner and outer bounds of the Frosting layer.} See text in Section~\ref{sec:shifts}.}
  \label{fig:shifts}
\end{figure}

As illustrated in Figure~\ref{fig:shifts}, to define this layer, we introduce two values $\innershift_i$ and $\outershift_i$ for each vertex $\vec{v}_i$ of the extracted base mesh $\calM$. This gives two surfaces with vertices $(\vec{v}_i+\innershift_i \vec{n}_i)_i$ and $(\vec{v}_i+\outershift_i \vec{n}_i)_i$ respectively, where $\vec{n}_i$ is the mesh normal at vertex $\vec{v}_i$. These two surfaces define the inner and outer bounds of the Frosting layer. Note that we do not have to build them explicitly as they directly depend on the base mesh and the $\innershift_i$'s and $\outershift_i$'s.

To find good values for the $\innershift_i$'s and $\outershift_i$'s, we initially tried using directly the unconstrained Gaussians, i.e., the Gaussians obtained before applying the regularization term from SuGaR.  Unfortunately, without regularization, Gaussian Splatting tends to retrieve a thick layer of Gaussians even for ``non-fuzzy'' surfaces, which would result in excessively large values for $\innershift_i$ and $\outershift_i$.  Moreover, the unconstrained Gaussians generally contain many transparent floaters and other outlier Gaussians. Such Gaussians could also bias the shifts toward unnecessarily large values. On the other hand, using only the regularized Gaussians to setup the $\innershift_i$'s and $\outershift_i$'s could miss fuzzy areas since these Gaussians are made flatter by the regularization.

Our solution is thus to consider both the unconstrained and the regularized Gaussians.  More exactly, we estimate the Frosting thickness from the thickness of the unconstrained Gaussians by looking for their isosurfaces, BUT, to make sure we consider the isosurfaces close to the scene surface, we search for the isosurfaces close to the regularized Gaussians: Even under the influence of the regularization term from SuGaR, Gaussians do not align well with the geometry around surfaces with fuzzy details.  As a consequence, the local thickness of the regularized Gaussians is a cue on how fuzzy the material is.

Figure~\ref{fig:shifts}  illustrates what we do to fix the $\innershift_i$'s and $\outershift_i$'s. To restrict the search, we define a first interval $I_i = [-3\sigma_i, 3\sigma_i]$ for each vertex $\vec{v}_i$, where $\sigma_i$ is the standard deviation in the direction of $\vec{n}_i$ of the regularized Gaussian the closest to $\vec{v}_i$. $I_i$ is the confidence interval for the 99.7 confidence level of the 1D Gaussian function of $t$ along the normal. Fuzzy parts result in general in large $I_i$. We could use the $I_i$'s to restrict the search for the isosurfaces of the unconstrained Gaussians. A more reliable search interval $J_i$ is obtained by looking for the isosurfaces of the regularized Gaussians along  $\vec{n}_i$ within $I_i$:
\begin{equation}
    \propinnershift_i = \inf ( T ) \text{ , }
    \propoutershift_i = \sup ( T ) \text{ , with }
    T = \left\{ t\in I_i \>\> | \>\> d_r(\vec{v}_i+t\vec{n}_i) \geq \lambda \right\} \> ,
    \label{eq:proposal-shifts}
\end{equation}
where $d_r$ is the density function as defined in Eq.~\eqref{eq:gaussian_splatting_density} for the regularized Gaussians.  In practice, we use an isosurface level~$\lambda = 0.01$, i.e., close to zero.
We use $ \propinnershift_i$ and $\propoutershift_i$ to define interval $J_i$: $J_i = \left[ \epsilon^\Mid_i - k\epsilon^\half_i, \epsilon^\Mid_i + k\epsilon^\half_i \right]$, with $\epsilon^\Mid_i = (\propinnershift + \propoutershift)/2$ and $\epsilon^\half_i = (\propoutershift - \propinnershift)/2$. We take $k=3$ as it gives an interval large enough to include most of the unconstrained Gaussians while rejecting the outlier Gaussians. Finally, we can compute the inner and outer shifts~$\innershift_i$ and $\outershift_i$ as:
\begin{equation}
    \innershift_i = \inf ( V ) \text{ , }
    \outershift_i = \sup ( V ) \text{ , with }
    V = \left\{ t\in J_i \>\> | \>\> d_u(\vec{v}_i+t\vec{n}_i) \geq \lambda \right\} \> .
    \label{eq:shifts}
\end{equation}

\begin{figure}[tb]
  \centering
  \begin{subfigure}{0.155\linewidth}
  \includegraphics[width=\linewidth]{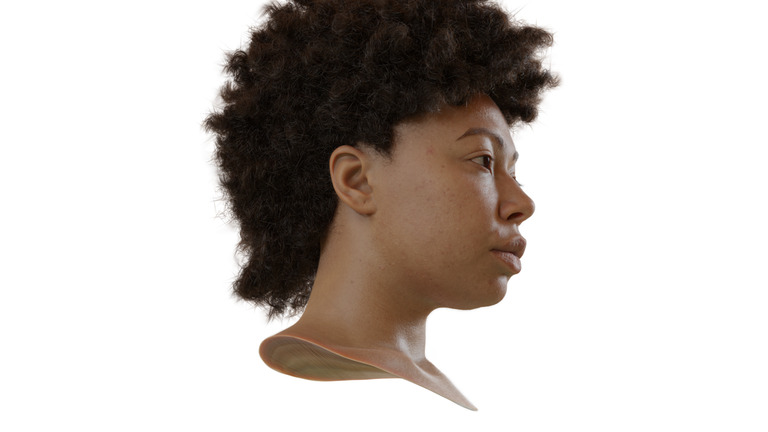}
  \includegraphics[width=\linewidth]{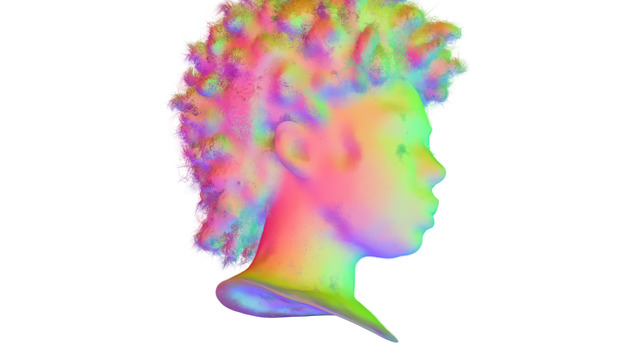}
  \includegraphics[width=\linewidth]{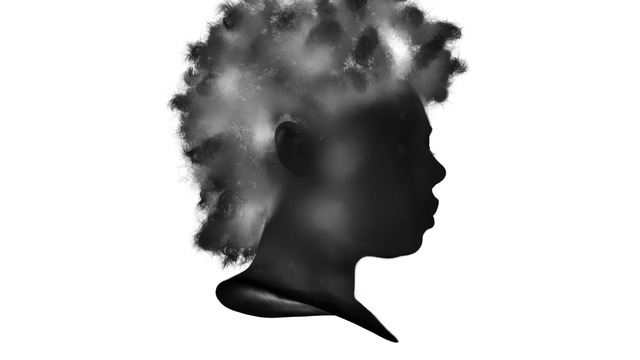}
  \end{subfigure}
  \hfill
  \begin{subfigure}{0.155\linewidth}
  \includegraphics[width=\linewidth]{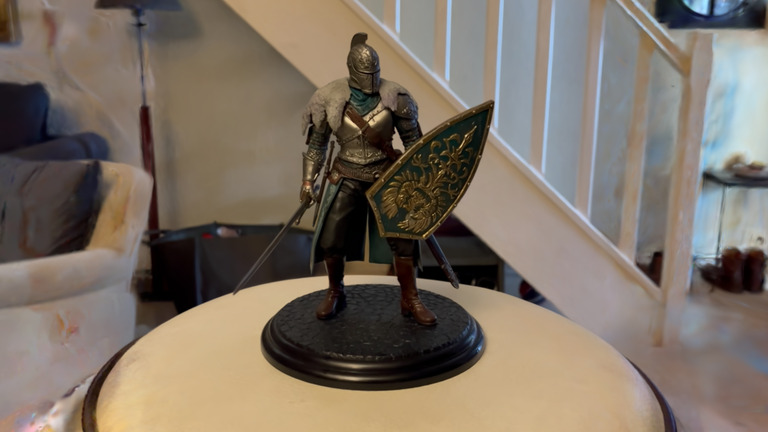}
  \includegraphics[width=\linewidth]{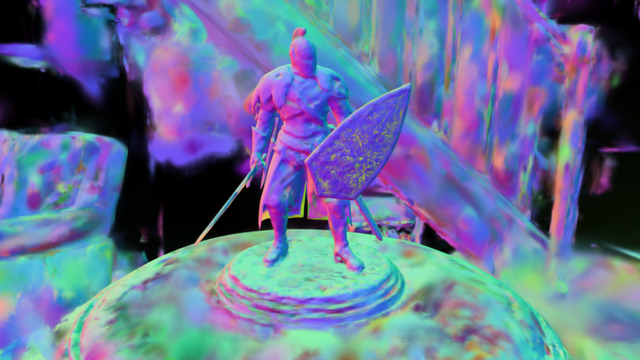}
  \includegraphics[width=\linewidth]{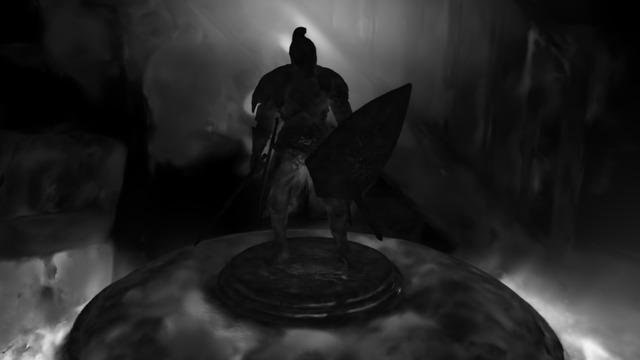}
  \end{subfigure}
  \hfill
  \begin{subfigure}{0.155\linewidth}
  \includegraphics[width=\linewidth]{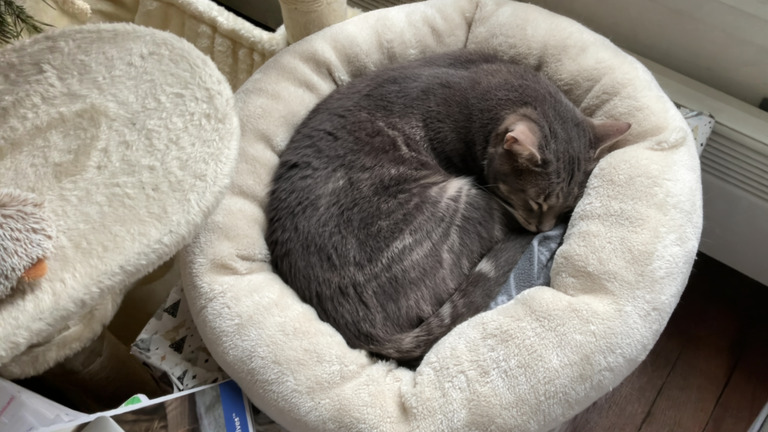}
  \includegraphics[width=\linewidth]{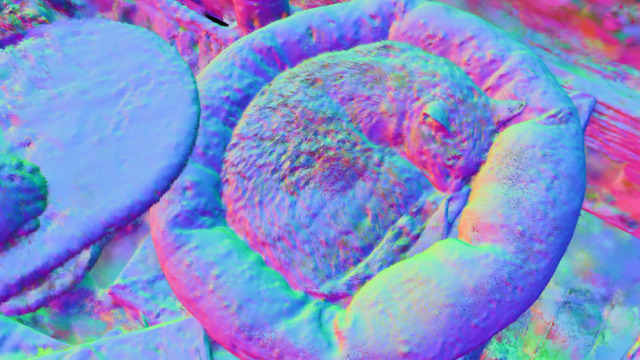}
  \includegraphics[width=\linewidth]{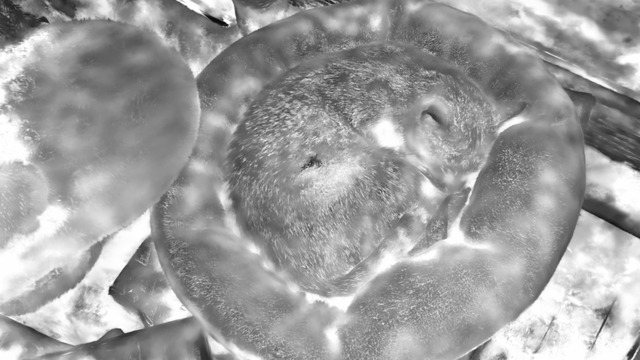}
  \end{subfigure}
  \hfill
  \begin{subfigure}{0.155\linewidth}
  \includegraphics[width=\linewidth]{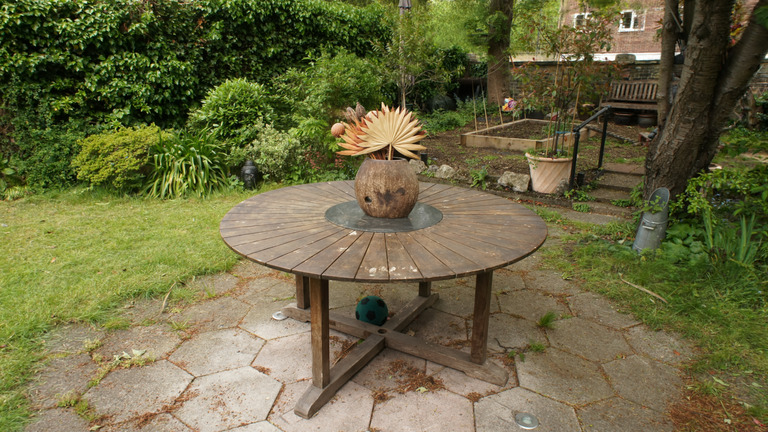}
  \includegraphics[width=\linewidth]{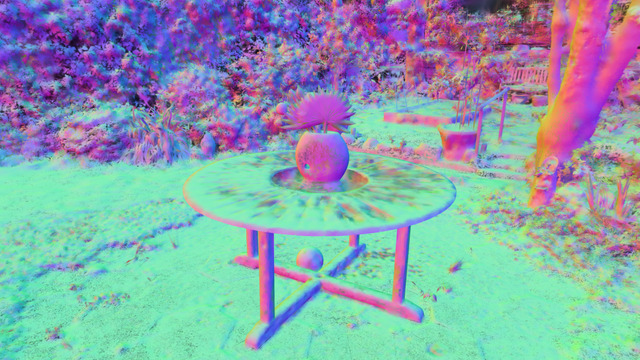}
  \includegraphics[width=\linewidth]{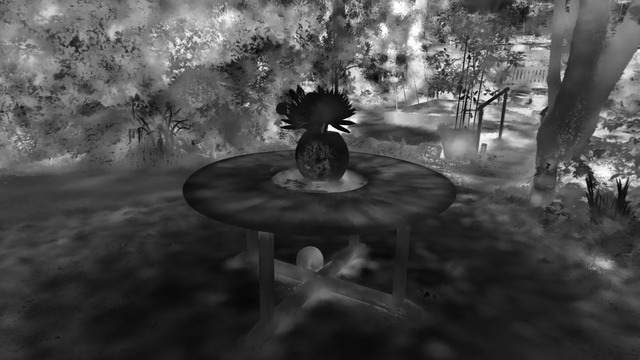}
  \end{subfigure}
  \hfill
  \begin{subfigure}{0.155\linewidth}
  \includegraphics[width=\linewidth]{images/renders/bicycle_rgb_52.jpg}
  \includegraphics[width=\linewidth]{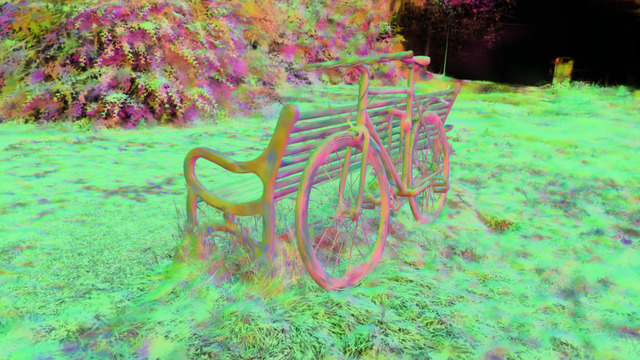}
  \includegraphics[width=\linewidth]{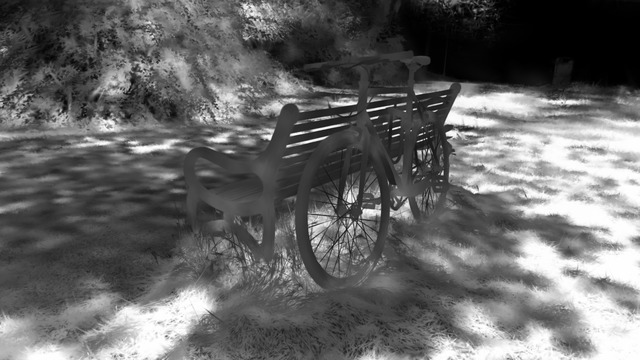}
  \end{subfigure}
  \caption{
  \textbf{Rendering complex scenes with Frosting.} First row: Renderings, Second row: recovered normal maps, Third row: estimated Frosting thickness. Note that the Frosting is thick on fuzzy materials such as the hair and the grass, as expected, and very thin on flat surfaces such as the table on the fourth column.
  }
  \label{fig:frosting-renders}
\end{figure}

\subsection{Frosting Optimization}

Once we constructed the outer and inner bounds of the Frosting layer, we initialize a densified set of Gaussians inside this layer and optimize them using 3DGS rendering loss as the unconstrained Gaussians. To make sure the Gaussians stay inside the frosting layer during optimization, we introduce a new parameterization of the Gaussians. Moreover, this parameterization will make possible to 
easily adjust the Gaussians' parameters when editing the scene.

\subsubsection{Parameterization.} Let us consider a triangular face of the base mesh $\calM$, with vertices denoted by~$\vec{v}_0, \vec{v}_1$, and~$\vec{v}_2$ and their corresponding normals~$\vec{n}_0, \vec{n}_1$, and~$\vec{n}_2$. After extracting inner and outer shifts from unconstrained Gaussians, we obtain six new vertices $(\vec{v}_i+\innershift_i \vec{n}_i)_{i=0,1,2}$ and $(\vec{v}_i+\outershift_i \vec{n}_i)_{i=0,1,2}$ that respectively belong to the inner and outer bounds of the frosting.
Specifically, these six vertices delimit an irregular triangular prism. We will refer to such polyhedrons as ``prismatic cells''. 
We parameterize the 3D mean~$\mu_g\in \IR^3$ of a Gaussian~$g \in \calG$ located inside a prismatic cell with a set of six barycentric coordinates split into two subsets~$(b_g^{(i)})_{i=0,1,2}$ and $(\beta_g^{(i)})_{i=0,1,2}$, such that 
\begin{equation}
    \mu_g = \sum_{i=0}^2 \left(
    b_g^{(i)} \left(\vec{v}_i+\outershift_i \vec{n}_i\right) +
    \beta_g^{(i)} \left(\vec{v}_i+\innershift_i \vec{n}_i\right)\right) \> ,
    \label{eq:barycentric-coordinates}
\end{equation}
with barycentric coordinates verifying $\sum_{i=0}^2 ( b_g^{(i)}+\beta_g^{(i)}) = 1$.
Using barycentric coordinates enforces Gaussians to stay inside their corresponding prismatic cell, and guarantees the stability of our representation during optimization.
In practice, we apply a softmax activation on the parameters to optimize to obtain barycentric coordinates that sum up to 1. 

\subsubsection{Initialization.} For a given budget $N$ of Gaussians provided by the user, we initialize $N$ Gaussians in the scene by sampling $N$ 3D centers $\mu_g$ in the frosting layer. Specifically, for sampling a single Gaussian, we first randomly select a prismatic cell with a probability proportional to its volume. Then, we sample random coordinates that sum up to 1. 
This sampling allows for allocating more Gaussians in areas with fuzzy and complex geometry, where more volumetric rendering is needed. However, flat parts  in the layer may also need a large number of Gaussians to recover texture details. Therefore, in practice, we instantiate $N/2$ Gaussians with uniform probabilities in the prismatic cells, and $N/2$ Gaussians with probabilities proportional to the volume of the cell.

We initialize the colors of the Gaussians with the color of the closest Gaussian in the unconstrained representation. However, we do not use the unconstrained Gaussians to initialize opacity, rotation, and scaling factors, as in practice, following the strategy from 3DGS~\cite{kerbl3Dgaussians} for these parameters provides better performance: 
We suppose the positions and configuration of the Gaussians inside the Frosting layer are already a good initialization, and resetting opacities, scaling factors and rotations helps Gaussians to take a fresh start, avoiding a potential local minimum encountered by previous unconstrained Gaussians.

Our representation allows for a much better control over the number of Gaussians than the original Gaussian Splatting densification process, as it is up to the user to decide on a number of Gaussians to instantiate in the frosting layer. These Gaussians will be spread in the entire frosting in a very efficient way, adapting to the need for volumetric rendering in the entire scene.

\subsubsection{Optimizing the Gaussian Frosting.} We reload the unconstrained Gaussians and apply our method for computing the inner and outer bounds of the Frosting. Then, for a given budget of $N$ Gaussians, we initialize $N$ Gaussians in the Frosting and optimize the representation while keeping the number of Gaussians constant. Note that compared to Vanilla 3DGS, this allows to control precisely the number of Gaussians.

\subsubsection{Editing, Deforming, and Animating the Frosting.} When deforming the base mesh, the positions of Gaussians automatically adjust in the frosting layer thanks to the use of the barycentric coordinates. 
To automatically adjust the rotation and scaling factors of the Gaussians, we propose a strategy different from the surface-based adjustment from SuGaR: In a given prismatic cell with center $\vec{c}$ and vertices $\vec{v}_i$ for $0\leq i<5$, we first estimate the local transformation at each vertex $\vec{v}_i$ by computing the rotation and rescaling of the vector $(c - \vec{v}_i)$. 
Then, we use the barycentric coordinates of a Gaussian $g$ to compute an average transformation at point $\mu_g$ from the transformation of all 6 vertices, and we adjust the rotation and scaling factors of $g$ by applying this average transformation. 
Please note that the spherical harmonics are also adjusted in practice, to ensure the consistency of the emitted radiance depending on the averaged rotation applied to the Gaussian.
We provide more details about this automatic adjustment of Gaussian parameters in the supplementary material.
\section{Experiments}

\begin{figure}[tb]
  \centering
  \begin{subfigure}{0.31\linewidth}
  \includegraphics[width=\linewidth]{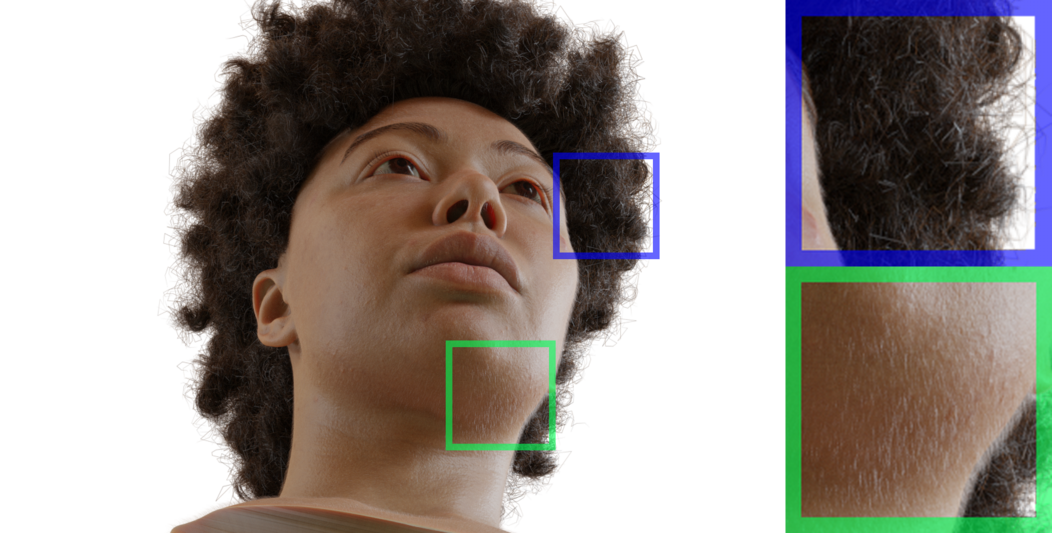}
  \includegraphics[width=\linewidth]{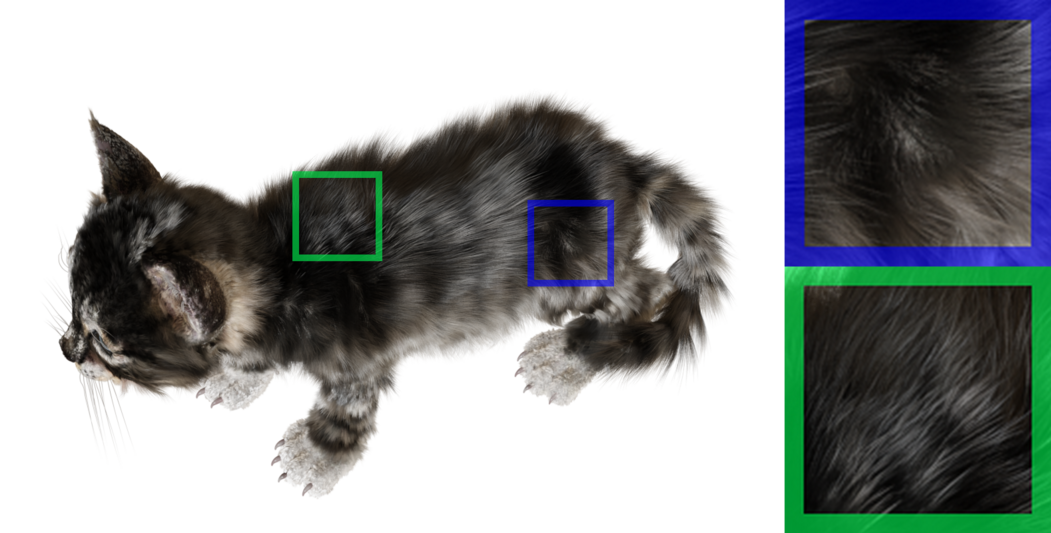}
  \includegraphics[width=\linewidth]{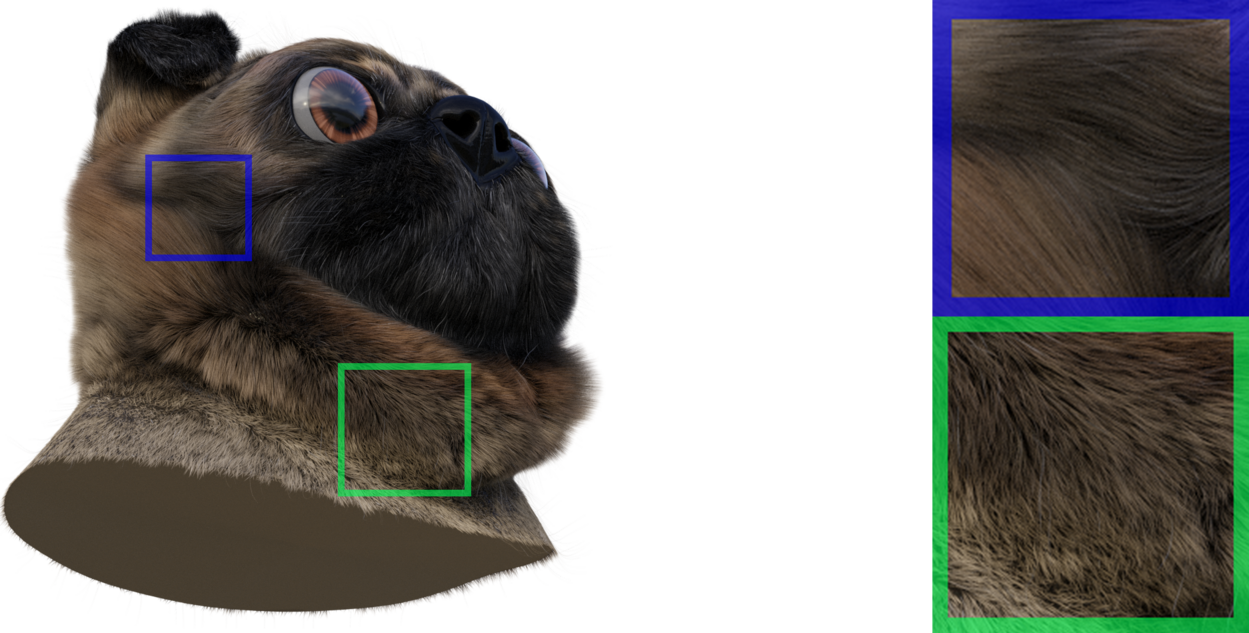}
  \caption{Ground Truth Image}
  \end{subfigure}
  \hfill
  \begin{subfigure}{0.31\linewidth}
  \includegraphics[width=\linewidth]{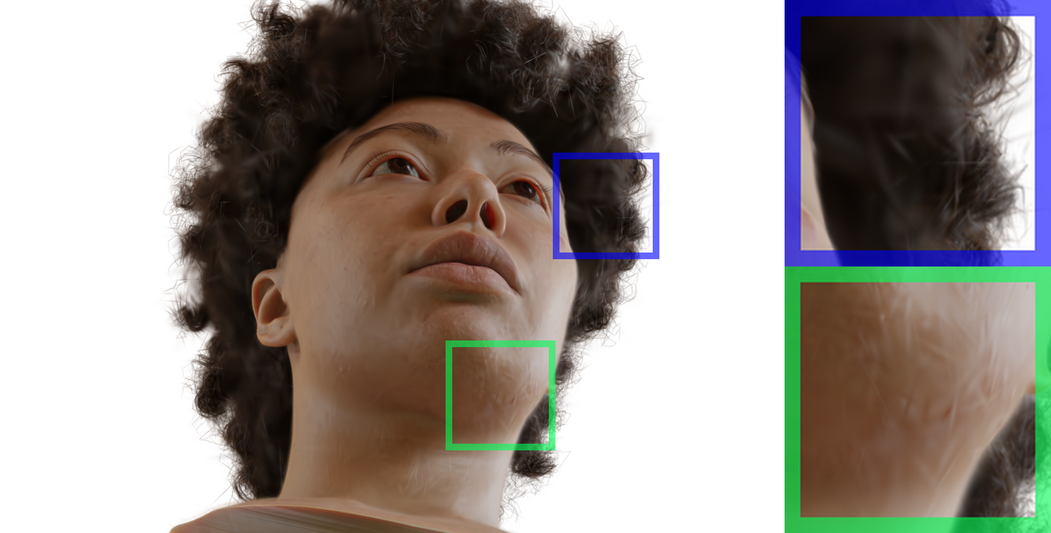}
  \includegraphics[width=\linewidth]{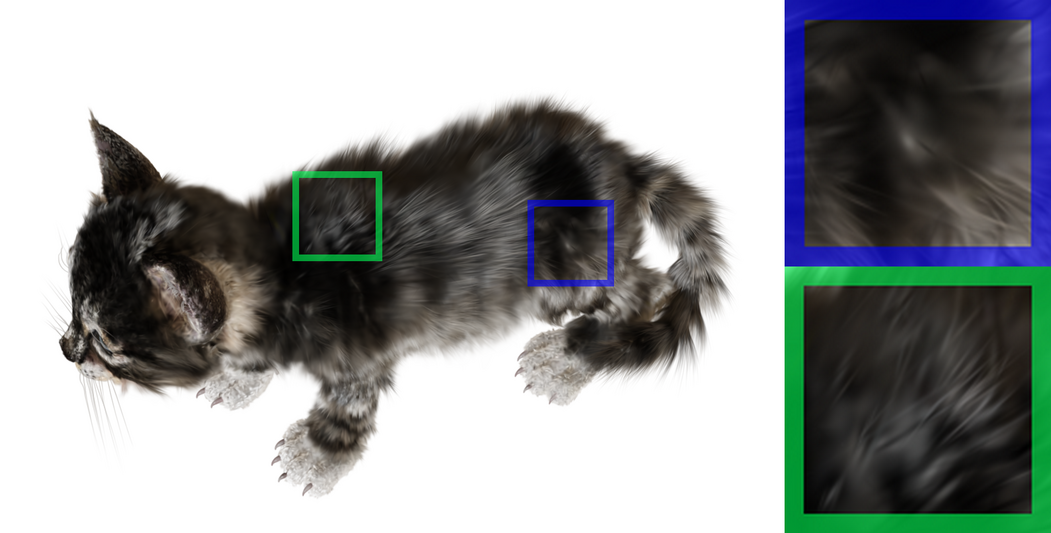}
  \includegraphics[width=\linewidth]{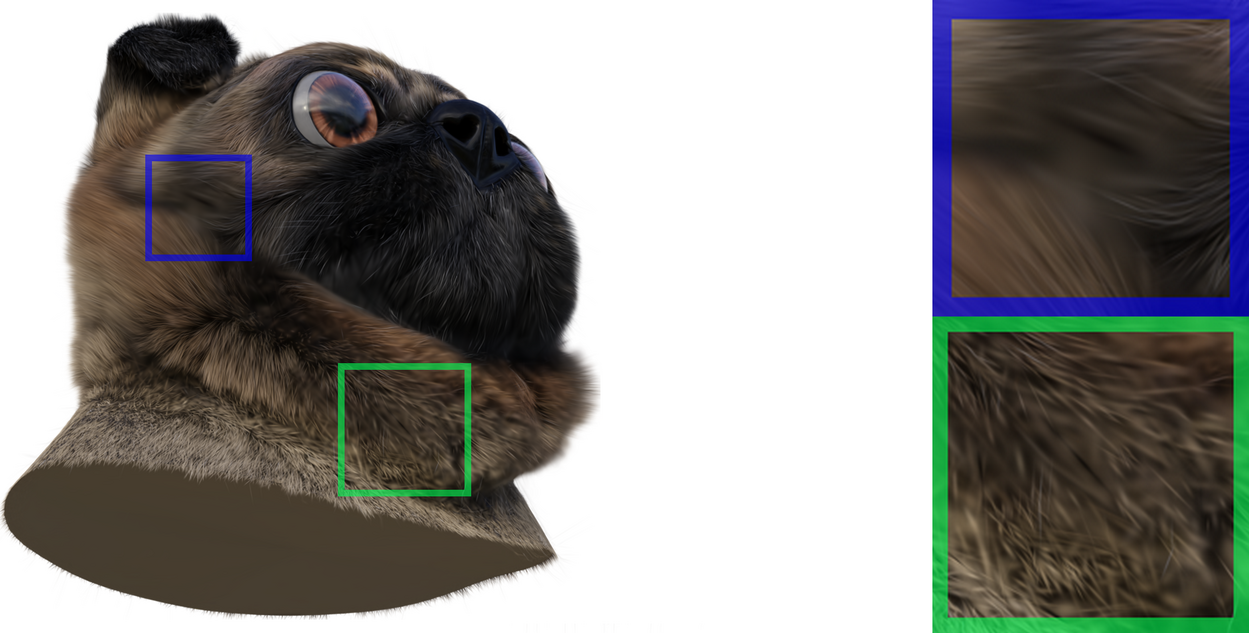}
  \caption{3DGS~\cite{kerbl3Dgaussians}}
  \end{subfigure}
  \hfill
  \begin{subfigure}{0.31\linewidth}
  \includegraphics[width=\linewidth]{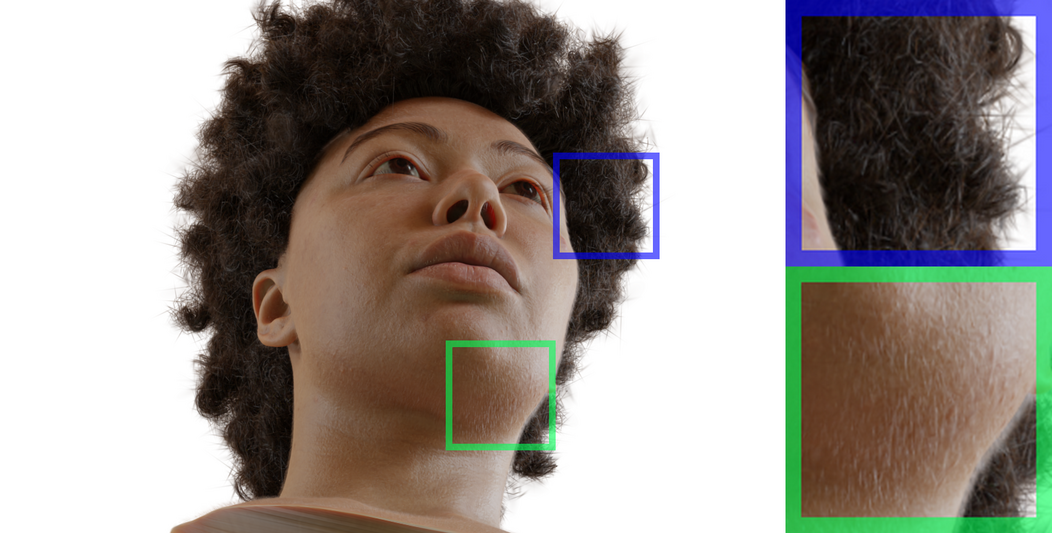}
  \includegraphics[width=\linewidth]{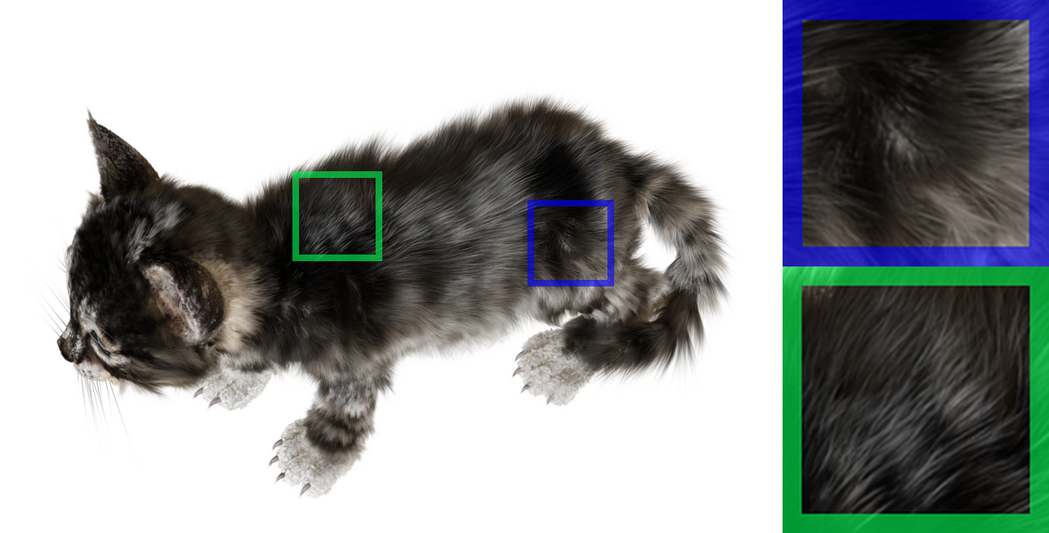}
  \includegraphics[width=\linewidth]{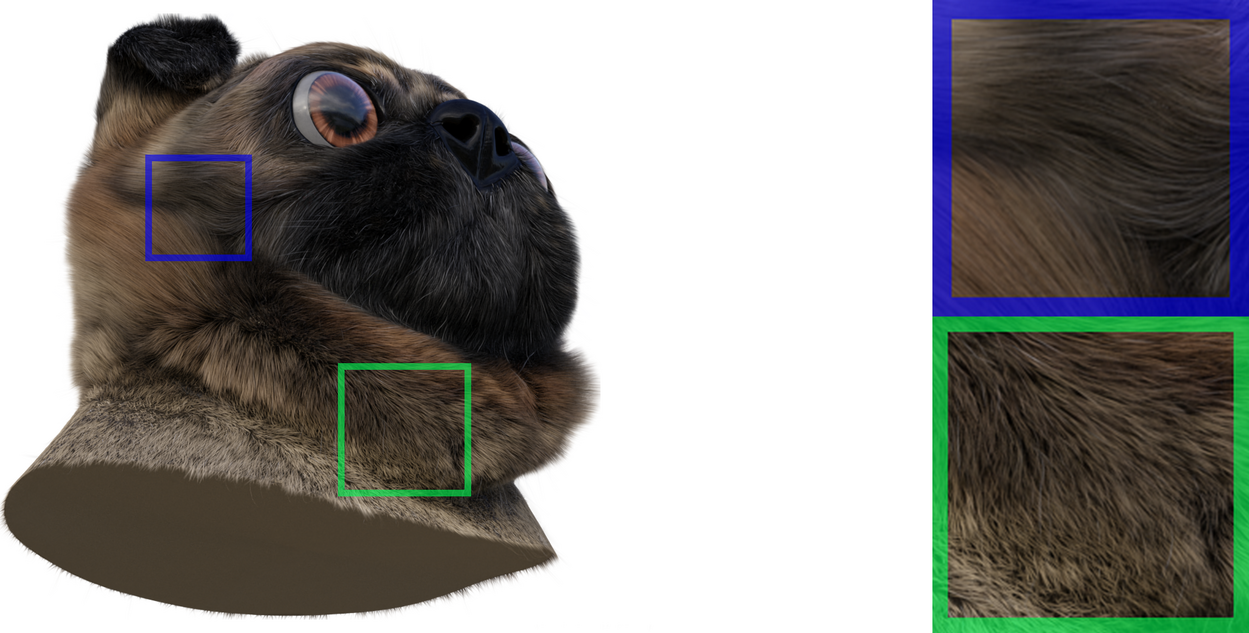}
  \caption*{Frosting (Ours)}
  \end{subfigure}
  \vspace{-0.3cm}
  \caption{
  \textbf{Close-up views of fuzzy materials from the Shelly dataset~\cite{wang-siggraphasia2023-adaptive-shells} reconstructed with vanilla Gaussian Splatting~\cite{kerbl3Dgaussians}~(center) and Frosting~(right).}
  }
  \label{fig:fuzzy-material-closeup}
\end{figure}

\subsection{Implementation Details}

We implemented our method with PyTorch~\cite{paszke-nips19-pytorch} and optimized the representations on a single GPU Nvidia Tesla V100~SXM2~32~Go.
Optimizing a full, editable Frosting model takes between 45 and 90 minutes on a single GPU, depending on the complexity of the scene. This optimization is much faster than the most similar approach to Frosting in the literature, namely Adaptive~Shells~\cite{wang-siggraphasia2023-adaptive-shells}, that requires 8~hours on a single GPU for a synthetic scene, and 1.7 times more iterations for a real scene.

\subsubsection{Extracting the surface mesh.} When reconstructing real scenes, we follow the approach from vanilla 3DGS~\cite{kerbl3Dgaussians} and first use COLMAP to estimate the camera poses and extract a  point cloud for initialization. For synthetic scenes with known camera poses, we just use a random point cloud for initialization. Then, we optimize an unconstrained Gaussian Splatting representation for 7,000 iterations. We save these Gaussians aside and apply the regularization term from SuGaR until iteration~15,000. We finally compute an optimal depth parameter $\bar{D}$ with $\gamma=100$ and extract a mesh from the regularized Gaussians by applying Poisson surface reconstruction as described in~\cite{guedon2023sugar}.

\subsubsection{Optimizing the Gaussian Frosting.} Given a budget of $N$ Gaussians, we initialize $N$ Gaussians in the Frosting layer and optimize them for 15,000 additional iterations, which gives a total of 30,000 iterations, similarly to 3DGS~\cite{kerbl3Dgaussians}. Vanilla 3DGS optimization generally produces between 1 and 5 million Gaussians. In practice, we use $N$=5~million for real scenes and $N$=2~million for synthetic scenes.

\subsection{Real-Time Rendering in Complex Scenes}

To evaluate the quality of Frosting's rendering, we compute the standard metrics PSNR, SSIM and LPIPS~\cite{zhang-2018-cvpr-lpips} and compare to several baselines, some of them focusing only on Novel View Synthesis~\cite{mildenhall2020nerf,wang2021neus,barron2021mipnerf,barron2022mipnerf360,mueller2022instantngp,yu_and_fridovichkeil2021plenoxels,kerbl3Dgaussians} and others relying on an editable representation~\cite{chen2022mobilenerf,rakotosaona2023nerfmeshing,yariv-2023-bakedsdf,reiser2024binaryopacitygrid,wang-siggraphasia2023-adaptive-shells,guedon2023sugar}, just like Frosting. We compute metrics on several challenging datasets containing synthetic and real scenes.\\

\noindent
\textbf{Shelly.} We first compare Frosting to state-of-the-art methods on the dataset Shelly introduced in Adaptive~Shells~\cite{wang-siggraphasia2023-adaptive-shells}. Shelly includes six synthetic scenes with challenging fuzzy materials that surface-based approaches struggle to reconstruct accurately. As we show in Table~\ref{tab:nvsmetrics_shelly} and Figure~\ref{fig:fuzzy-material-closeup}, Frosting outperforms every other methods for all three metrics. Frosting even outperforms with a wide margin vanilla Gaussian Splatting, which is free from any surface constraints and only focuses on optimizing the rendering quality. Indeed, the sampling of Gaussians inside the Frosting layer provides a much more efficient densification of Gaussians than the strategy proposed in 3DGS~\cite{kerbl3Dgaussians}, targeting the challenging fuzzy areas close to the surface and allocating more Gaussians where volumetric rendering is needed.\\

\noindent
\textbf{NeRFSynthetic.} Table~\ref{tab:nvsmetrics_shelly} provides a comparison on the NeRFSynthetic data\-set~\cite{mildenhall2020nerf}, which consists in eight synthetic scenes. Frosting performs the best among the editable methods, surpassing Su\-GaR~\cite{guedon2023sugar}, and achieves results on par with vanilla 3DGS and other radiance field methods.\\

\begin{table}
   \caption{
   \antoine{
   \textbf{Quantitative evaluation of rendering quality on the synthetic datasets \emph{Shelly}~\cite{wang-siggraphasia2023-adaptive-shells} and \emph{NeRFSynthetic}~\cite{mildenhall2020nerf}.} Frosting is the best among all methods, outperforming even non-editable models that only focus on rendering. Contrary to an unconstrained 3D Gaussian Splatting~\cite{kerbl3Dgaussians}, our representation allows for densifying Gaussians more efficiently by targeting challenging and fuzzy areas.}}
  \label{tab:nvsmetrics_shelly}
  \centering
  {\scriptsize
  \begin{tabular}{@{}lccccccc@{}}
    \toprule
     \multicolumn{1}{c}{} & \multicolumn{3}{c}{\emph{Shelly}} & \multicolumn{3}{c}{\emph{NeRFSynthetic}} \\
     \cmidrule(r){2-4} \cmidrule(r){5-7}
      & PSNR $\uparrow$ & SSIM $\uparrow$ & LPIPS $\downarrow$ & PSNR $\uparrow$ & SSIM $\uparrow$ & LPIPS $\downarrow$ & \\
     %
    \midrule
    NeRF~\cite{mildenhall2020nerf} & 31.27 & 0.893 & 0.157 & 31.01 & 0.947 & 0.081 & \\
    NeuS~\cite{wang2021neus} & 29.98 & 0.893 & 0.158 & -- & -- & -- & \\
    Mip-NeRF~\cite{barron2021mipnerf}  & 32.59 & 0.899 & 0.148 & \cellcolor{yellow!25}33.09 & 0.961 & 0.043 & \\
    I-NGP~\cite{mueller2022instantngp} & 33.22 & 0.922 & 0.125 & \cellcolor{orange!25}33.18 & -- & -- & \\
    3DGS~\cite{kerbl3Dgaussians}  & \cellcolor{orange!25}37.66 & \cellcolor{orange!25}0.958 & \cellcolor{orange!25}0.066 & \cellcolor{red!25}\textbf{33.32} & \cellcolor{red!25}\textbf{0.970} & \cellcolor{orange!25}0.030 & \\
    \midrule
    MobileNeRF~\cite{chen2022mobilenerf} & 31.62 & 0.911 & 0.129 & 30.90 & 0.947 & 0.062 & \\
    Adaptive Shells~\cite{wang-siggraphasia2023-adaptive-shells}  & 36.02 & \cellcolor{yellow!25}0.954 & 0.079 & 31.84 & 0.957 & 0.056 & \\
    SuGaR\cite{guedon2023sugar}  & \cellcolor{yellow!25}36.33 & \cellcolor{yellow!25}0.954 & \cellcolor{yellow!25}0.059 & 32.40 & \cellcolor{yellow!25}0.964 & \cellcolor{yellow!25}0.033 & \\
    Frosting (Ours)  & \cellcolor{red!25}\textbf{39.84} & \cellcolor{red!25}\textbf{0.977} & \cellcolor{red!25}\textbf{0.033} & 33.03 & \cellcolor{orange!25}0.967 & \cellcolor{red!25}\textbf{0.029} & \\
    \bottomrule
  \end{tabular}
  }
\end{table}

\noindent
\textbf{Mip-NeRF~360.} We also compare Frosting to state-of-the-art approaches on the real scenes from the Mip-NeRF~360 dataset~\cite{barron2022mipnerf360}. This dataset contains images from seven challenging real scenes, but was captured with ideal lighting condition and provides really good camera calibration data and initial SfM~points. Results are available in Table~\ref{tab:nvsmetrics_mipnerf360} and Figure~\ref{fig:frosting-renders}. Frosting reaches the best performance among all editable methods, and obtains worse but competitive results compared to vanilla Gaussian Splatting. When Gaussian Splatting is given a very good initialization with a large amount of SfM points, the benefits from the Gaussian Frosting densification are not as effective, and optimizing Gaussians without additional constraints as in 3DGS slightly improves performance.\\

\noindent
\textbf{Additional real scenes.} We finally compare Frosting to the baselines with captures of real scenes that present variations in exposure or white balance. To this end, we follow the approach from 3DGS~\cite{kerbl3Dgaussians} and select the same two subsets of two scenes from \emph{Tanks\&Temples} (\emph{Truck} and \emph{Train}) and \emph{Deep~Blending}~(\emph{Playroom} and \emph{Dr.~Johnson}). We also evaluate a few methods on a custom dataset that consists of four casual captures made with a smartphone (we call these scenes \emph{SleepyCat}, \emph{Buzz}, \emph{RedPanda}, and \emph{Knight}), illustrated in Figures~\ref{fig:teaser} and~\ref{fig:frosting-renders}. Results are available in Table~\ref{tab:nvsmetrics_tandtdb}. In these more realistic scenarios, Frosting achieves once again similar or better performance than unconstrained Gaussian Splatting even though it is an editable representation that relies on a single, animatable mesh.

\begin{table}
   \caption{\textbf{Quantitative evaluation of rendering quality on the Mip-NeRF~360 dataset~\cite{barron2022mipnerf360}.} \antoine{Frosting is best among the methods that recover an editable Radiance Field with explicit meshes, and achieves performance comparable to NeRF methods and vanilla 3D Gaussian Splatting.} }
  \label{tab:nvsmetrics_mipnerf360}
  \centering
  {\scriptsize
  \begin{tabular}{@{}lcccccccccc@{}}
    \toprule
     \multicolumn{1}{c}{} & \multicolumn{3}{c}{Indoor scenes} & \multicolumn{3}{c}{Outdoor scenes} & \multicolumn{3}{c}{Average on all scenes} \\
     \cmidrule(r){2-4} \cmidrule(r){5-7} \cmidrule(r){8-10}
      & PSNR $\uparrow$ & SSIM $\uparrow$ & LPIPS $\downarrow$ & PSNR $\uparrow$ & SSIM $\uparrow$ & LPIPS $\downarrow$ & PSNR $\uparrow$ & SSIM $\uparrow$ & LPIPS $\downarrow$ \\
    \midrule
    \multicolumn{10}{l}{\textbf{No mesh (except Frosting)}} \\
    \midrule
    Plenoxels~\cite{yu_and_fridovichkeil2021plenoxels} & 24.83 & 0.766 & 0.426 & 22.02 & 0.542 & 0.465 & 23.62 & 0.670 & 0.443 \\
    INGP-Base~\cite{mueller2022instantngp} & 28.65 & 0.840 & 0.281 & 23.47 & 0.571 & 0.416 & 26.43 & 0.725 & 0.339 \\
    INGP-Big~\cite{mueller2022instantngp} & 29.14 & 0.863 & 0.242 & 23.57 & 0.602 & 0.375 & 26.75 & 0.751 & 0.299 \\
    Mip-NeRF 360~\cite{barron2022mipnerf360} & \cellcolor{red!25}\textbf{31.58} & \cellcolor{yellow!25}0.914 & \cellcolor{red!25}\textbf{0.182} & \cellcolor{orange!25}25.79 & \cellcolor{yellow!25}0.746 & \cellcolor{yellow!25}0.247 & \cellcolor{red!25}\textbf{29.09} & \cellcolor{yellow!25}0.842 & \cellcolor{yellow!25}0.210 \\
    3DGS~\cite{kerbl3Dgaussians} & \cellcolor{yellow!25}30.41 & \cellcolor{orange!25}0.920 & \cellcolor{orange!25}0.189 & \cellcolor{red!25}\textbf{26.40} & \cellcolor{red!25}\textbf{0.805} & \cellcolor{red!25}\textbf{0.173} & \cellcolor{orange!25}28.69 & \cellcolor{red!25}\textbf{0.870} & \cellcolor{red!25}\textbf{0.182} \\
    Frosting (Ours) & \cellcolor{orange!25}30.49 & \cellcolor{red!25}\textbf{0.925} & \cellcolor{yellow!25}0.190 & \cellcolor{yellow!25}25.57 & \cellcolor{orange!25}0.765 & \cellcolor{orange!25}0.225 & \cellcolor{yellow!25}28.38 & \cellcolor{orange!25}0.856 & \cellcolor{orange!25}0.205\\
    \midrule
    \multicolumn{10}{l}{\textbf{With mesh}} \\
    \midrule
    MobileNeRF~\cite{chen2022mobilenerf} & 25.74 & 0.757 & 0.453 & 22.90 & 0.524 & 0.463 & 24.52 & 0.657 & 0.457 \\
    NeRFMeshing~\cite{rakotosaona2023nerfmeshing} & 23.83 & -- & -- & 22.23 & -- & -- & 23.15 & -- & -- \\
    BakedSDF~\cite{yariv-2023-bakedsdf} & 27.20 & 0.845 & 0.300 & \cellcolor{yellow!25}23.40 & 0.577 & \cellcolor{yellow!25}0.351 & 25.57 & 0.730 & \cellcolor{yellow!25}0.321 \\
    B.O. Grids~\cite{reiser2024binaryopacitygrid} & 27.71 & \cellcolor{yellow!25}0.873 & \cellcolor{yellow!25}0.227 & -- & -- & -- & -- & -- & -- \\
    Adaptive Shells~\cite{wang-siggraphasia2023-adaptive-shells} & \cellcolor{yellow!25}29.19 & 0.872 & 0.285 & 23.17  & \cellcolor{yellow!25}0.606 & 0.389 & \cellcolor{yellow!25}26.61 & \cellcolor{yellow!25}0.758 & 0.330 \\
    SuGaR\cite{guedon2023sugar} & \cellcolor{orange!25}29.43 & \cellcolor{orange!25}0.910 & \cellcolor{orange!25}0.216 & \cellcolor{orange!25}24.40 & \cellcolor{orange!25}0.699 & \cellcolor{orange!25}0.301 & \cellcolor{orange!25}27.27 & \cellcolor{orange!25}0.820 & \cellcolor{orange!25}0.253 \\
    Frosting (Ours) & \cellcolor{red!25}\textbf{30.49} & \cellcolor{red!25}\textbf{0.925} & \cellcolor{red!25}\textbf{0.190} & \cellcolor{red!25}\textbf{25.57} & \cellcolor{red!25}\textbf{0.765} & \cellcolor{red!25}\textbf{0.225} & \cellcolor{red!25}\textbf{28.38} & \cellcolor{red!25}\textbf{0.856} & \cellcolor{red!25}\textbf{0.205}\\
    \bottomrule
  \end{tabular}
  }
\end{table}
\begin{table}
   \caption{
   \antoine{
   \textbf{Quantitative evaluation of rendering quality on real scenes from \emph{Tanks\&Temples}~\cite{knapitsch-2017-tanksandtemples}, \emph{Deep~Blending}~\cite{hedman-2018-deepblending} and our custom dataset.} Our representation performs the best among the surface-based methods, and achieves similar or better performance than unconstrained 3DGS and other non-editable methods.}}
  \label{tab:nvsmetrics_tandtdb}
  \centering
  {\scriptsize
  \begin{tabular}{@{}lcccccccccc@{}}
    \toprule
     \multicolumn{1}{c}{} & \multicolumn{3}{c}{\emph{Tanks\&Temples}~\cite{knapitsch-2017-tanksandtemples}} & \multicolumn{3}{c}{\emph{Deep~Blending}~\cite{hedman-2018-deepblending}} & \multicolumn{3}{c}{\emph{Custom dataset}} \\
     \cmidrule(r){2-4} \cmidrule(r){5-7} \cmidrule(r){8-10}
      & PSNR $\uparrow$ & SSIM $\uparrow$ & LPIPS $\downarrow$ & PSNR $\uparrow$ & SSIM $\uparrow$ & LPIPS $\downarrow$ & PSNR $\uparrow$ & SSIM $\uparrow$ & LPIPS $\downarrow$ & \\
    \midrule
    Plenoxels~\cite{yu_and_fridovichkeil2021plenoxels}  & 21.07 & 0.719 & 0.379 & 23.06 & 0.794 & 0.510 & -- & -- & -- &\\
    INGP-Base~\cite{mueller2022instantngp}  & 21.72 & 0.723 & 0.330 & 23.62 & 0.796 & 0.423 & -- & -- & -- & \\
    INGP-Big~\cite{mueller2022instantngp}  & 21.92 & 0.744 & 0.304 & 24.96 & 0.817 & 0.390 & -- & -- & -- & \\
    Mip-NeRF~360~\cite{barron2022mipnerf360}  & \cellcolor{yellow!25}22.22 & 0.758 & 0.257 & 29.40 & \cellcolor{orange!25}0.901 & \cellcolor{yellow!25}0.244 & -- & -- & -- & \\
    3DGS~\cite{kerbl3Dgaussians} & \cellcolor{red!25}23.14 & \cellcolor{red!25}0.841 & \cellcolor{orange!25}0.183 & \cellcolor{orange!25}29.41 & \cellcolor{red!25}0.903 & \cellcolor{orange!25}0.243 & \cellcolor{red!25}34.17 & \cellcolor{orange!25}0.944 & \cellcolor{orange!25}0.165 & \\
    \midrule
    SuGaR\cite{guedon2023sugar}  & 21.58 & \cellcolor{yellow!25}0.795 & \cellcolor{yellow!25}0.219 & \cellcolor{orange!25}29.41 & 0.893 & 0.267 & \cellcolor{yellow!25}32.05 & \cellcolor{yellow!25}0.930 & \cellcolor{yellow!25}0.180 & \\
    Frosting (Ours)  & \cellcolor{orange!25}23.13 & \cellcolor{orange!25}0.836 & \cellcolor{red!25}0.174 & \cellcolor{red!25}29.62 & \cellcolor{yellow!25}0.900 & \cellcolor{red!25}0.236 & \cellcolor{orange!25}33.82 & \cellcolor{red!25}0.945 & \cellcolor{red!25}0.149 & \\
    \bottomrule
  \end{tabular}
  }
\end{table}

\subsection{Editing, Compositing, and Animating Gaussian Frosting}

\begin{figure}[tb]
  \centering
  \begin{subfigure}{0.31\linewidth}
 \includegraphics[trim={4cm 0 5cm 0},clip,width=\linewidth]{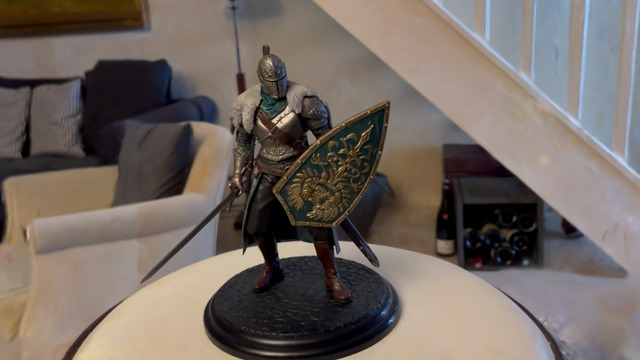}
 \caption{Original pose}
  \end{subfigure}
  \hfill
  \begin{subfigure}{0.31\linewidth}
  \includegraphics[trim={4cm 0 5cm 0},clip,width=\linewidth]{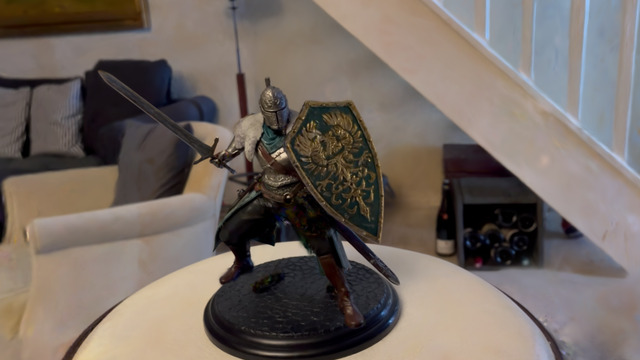}
  \caption{Edited pose}
  \end{subfigure}
  \hfill
  \begin{subfigure}{0.31\linewidth}
  \includegraphics[trim={4cm 0 5cm 0},clip,width=\linewidth]{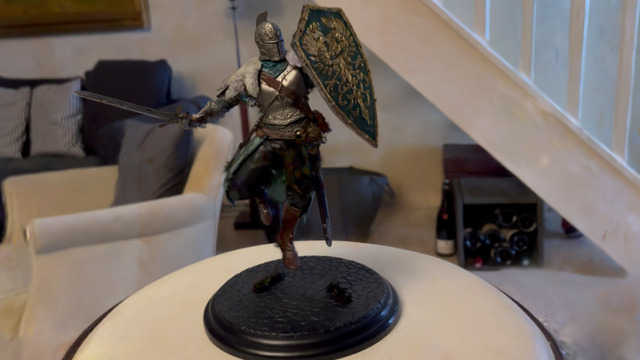}
  \caption{Edited pose}
  \end{subfigure}
  \caption{
  \textbf{Examples of animation with Frosting.} We were able to animate the sculpture in the left image using the rigging tool in Blender.
  }
  \label{fig:edition-examples}
\end{figure}

As shown in Figure~\ref{fig:sugar-comparison}, Figure~\ref{fig:scene-composition} and Figure~\ref{fig:edition-examples}, our Frosting representation automatically adapts when editing, rescaling, deforming, combining or animating base meshes.
Frosting offers editing, composition and animation capabilities similar to surface-based approaches like SuGaR~\cite{guedon2023sugar}, but achieves much better performance thanks to its frosting layer with variable thickness that adapts to the volumetric effects and fuzzy materials in the scene.

\subsection{Ablation Study: Octree Depth}
\begin{table}
   \caption{
   \textbf{Ablation for two different depth computation methods used for the octree in Poisson surface reconstruction.} We compare the rendering performance between using a predefined depth with a high value as in~\cite{guedon2023sugar} and our automatically computed depth. Our technique systematically selects an optimal depth depending on the complexity of the scene, avoiding artifacts in the mesh, and resulting in equivalent or better rendering performance with a much smaller average number of triangles.
   }
  \label{tab:ablation-octree-depth}
  \centering
  {\scriptsize
  \begin{tabular}{@{}lccccccccc@{}}
    \toprule
     \multicolumn{1}{c}{} & \multicolumn{4}{c}{\emph{NeRFSynthetic}} & \multicolumn{4}{c}{\emph{Shelly}} \\
     \cmidrule(r){2-5} \cmidrule(r){6-9}
      & PSNR $\uparrow$ & SSIM $\uparrow$ & LPIPS $\downarrow$ & $N_{tri}$ $\downarrow$  & PSNR $\uparrow$ & SSIM $\uparrow$ & LPIPS $\downarrow$ & $N_{tri}$ $\downarrow$ & \\
    \midrule
    Predefined depth $D$ = 10~\cite{guedon2023sugar} & 31.63 & 0.959 & 0.041 & >1 M & \cellcolor{red!25}\textbf{39.85} & 0.975 & 0.035 & 939 K &\\
    Depth $\bar{D} \leq 10$ & \cellcolor{red!25}\textbf{33.03} & \cellcolor{red!25}\textbf{0.967} & \cellcolor{red!25}\textbf{0.029} & \cellcolor{red!25}\textbf{863 K} & 39.84 & \cellcolor{red!25}\textbf{0.977} & \cellcolor{red!25}\textbf{0.033} & \cellcolor{red!25}\textbf{203 K} &\\
    \bottomrule
  \end{tabular}
  }
\end{table}

To demonstrate how our technique for automatically computing the optimal octree depth $\bar{D}$ for Poisson reconstruction improves the performance of Frosting, we provide in Table~\ref{tab:ablation-octree-depth} a comparison in rendering performance between our full model, and a version of Frosting that uses the same predefined depth parameter as SuGaR. This technique results in equivalent or better rendering performance with much fewer triangles.

\subsection{Ablation Study: Thickness of the Frosting Layer}
\begin{table}
   \caption{
   \textbf{Comparing different strategies for computing the thickness of the Frosting layer.} We compare the rendering performance (PSNR $\uparrow$) in synthetic and real scenes depending on how we compute and refine the thickness of the Frosting layer. Specifically, we first show that using an adaptive thickness improves performance over a constant thickness. Even though using a large constant thickness improves performance in scenes with very fuzzy materials like Shelly~\cite{wang-siggraphasia2023-adaptive-shells}, this lowers performance in scenes with flat surfaces and generates artifacts when editing the scene, as shown in Figure~\ref{fig:artifacts}. By contrast, our method adapts automatically to the type of surfaces. We also demonstrate that refining the thickness using the unconstrained Gaussians is necessary to achieve top performance.}
  \label{tab:ablation-proposal-intervals}
  \centering
  {\scriptsize
  \begin{tabular}{@{}lccccc@{}}
    \toprule
     \multicolumn{1}{c}{} & \multicolumn{1}{c}{\emph{\> Shelly \>}} & \multicolumn{3}{c}{\emph{Mip-NeRF~360}} \\
    \cmidrule(r){2-2} \cmidrule(r){3-5}
     & Average & Indoor & Outdoor & Average & \\
     %
    \midrule
    Constant thickness (Small)  & 39.03 & \cellcolor{yellow!25}30.36 & 25.50 & \cellcolor{yellow!25}28.28 &  \\
    Constant thickness (Medium)  & \cellcolor{yellow!25}39.67 & 30.19 & \cellcolor{yellow!25}25.54 & 28.20 &  \\
    Constant thickness (Large) & \cellcolor{red!25}\textbf{40.00} & 30.06 & 25.48 & 28.10 &\\
    Using Regularized Gaussians only ($\delta^{\text{in/out}}=\epsilon^{\text{in/out}}$) & 39.34 & \cellcolor{orange!25}30.42 & \cellcolor{red!25}\textbf{25.57} & \cellcolor{orange!25}28.34 &  \\
     Using Regularized and Unconstrained Gaussians (Full method)  & \cellcolor{orange!25}39.84 & \cellcolor{red!25}\textbf{30.49} & \cellcolor{red!25}\textbf{25.57} & \cellcolor{red!25}\textbf{28.38} & \\
    \bottomrule
  \end{tabular}
  }
\end{table}
\begin{figure}[tb]
  \centering
  \begin{subfigure}{0.31\linewidth}
 \includegraphics[width=\linewidth]{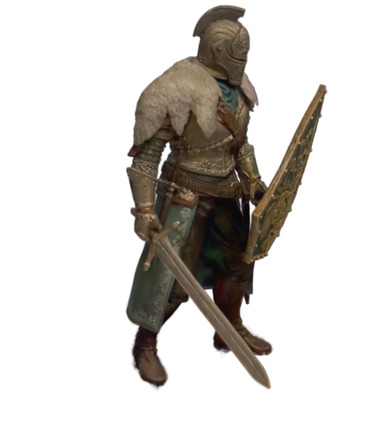}
 \caption{\tiny Original pose}
  \end{subfigure}
  \hfill
  \begin{subfigure}{0.31\linewidth}
  \includegraphics[width=\linewidth]{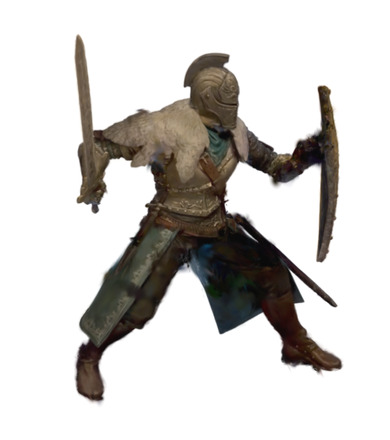}
  \caption{\tiny Adaptive thickness (ours)}
  \end{subfigure}
  \hfill
  \begin{subfigure}{0.31\linewidth}
  \includegraphics[width=\linewidth]{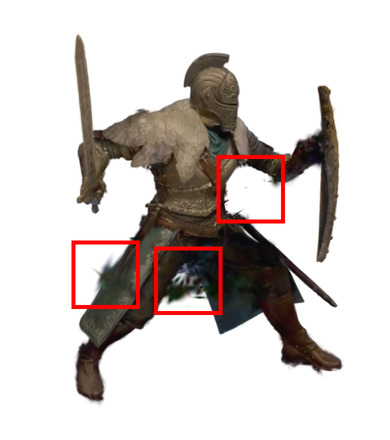}
  \caption{\tiny Constant thickness (baseline)}
  \end{subfigure}
  \caption{
  \textbf{Comparison with a constant thickness.} Our strategy to compute an adaptive thickness for Frosting is essential to maintain optimal performance while avoiding artifacts when editing the scene. As shown in the right image, using a constant thickness may produce artifacts when animating a character: When using a constant, large thickness in this scene, Gaussians located near the right knee of the knight participate in reconstructing the right hand, which produces artifacts when moving the hand.
  }
  \label{fig:artifacts}
\end{figure}

We also provide in Table~\ref{tab:ablation-proposal-intervals} an ablation study comparing different strategies for computing the thickness of the Frosting layer. 
Specifically, we first evaluate the rendering performance of a Frosting layer with constant thickness. We repeat the experiment for small, medium and large thickness values, using different quantiles of our inner and outer shifts~$\innershift$ and $\outershift$ for computing the constant thickness.
We show that using an adaptive thickness improves performance over a constant thickness, as (a) some fuzzy materials need a thicker frosting to be accurately rendered, and (b) some flat surfaces are better rendered with a very thin frosting. 
As a consequence, even though using a large constant thickness improves performance in scenes with very fuzzy materials like Shelly~\cite{wang-siggraphasia2023-adaptive-shells}, it lowers performance in scenes with flat surfaces. Moreover, using an adaptive thickness rather than a constant thickness with a large value helps to greatly reduce artifacts, as we demonstrate in Figure~\ref{fig:artifacts}.

We also show that using unconstrained Gaussians to refine the thickness of the Frosting is necessary to achieve top performance. To this end, we skip the refinement process and evaluate the rendering performance of a Frosting layer with $\innershift=\propinnershift$ and $\outershift=\propoutershift$. This results in lower performance, as shown in Table~\ref{tab:ablation-proposal-intervals}.
\section{Conclusion}

We proposed a simple yet powerful surface representation with many advantages over current representations together with a method to extract it from images. One limitation of our implementation is the simple deformation model as it is piecewise linear. It should be however simple to replace it with a more sophisticated, physics-based deformation model. Another limitation is that our models are larger than vanilla Gaussian Splatting models since we have to include the barycentric coordinates and the mesh vertices.  Recent works about compressing 3DGS could help.

We believe that the Frosting representation can be useful beyond image-based rendering. It could for example be used in more general Computer Graphics applications to render complex materials in real-time.
\newpage

%
%
\bibliographystyle{splncs04}
\bibliography{arxiv}

\newpage
\centerline{\large Supplementary Material}

\vspace{0.3in}
\noindent
In this supplementary material, we provide the following elements:
\begin{itemize}
    \item A description of our method to improve the surface reconstruction from SuGaR~\cite{guedon2023sugar}.
    \item Additional details about our strategy to initialize the Frosting layer and automatically adjust Gaussians' parameters when deforming, editing, or animating our representation.
\end{itemize}
We also provide a \href{https://anttwo.github.io/frosting/}{\underline{video}} that offers an overview of the approach and showcases additional qualitative results. Specifically, the video demonstrates how Frosting can be used to edit, combine or animate Gaussian Splatting representations.

\setcounter{section}{+6}
\section{Improving surface reconstruction}


We improve the surface reconstruction method from SuGaR by proposing a way to automatically adjust the hyperparameter of the Poisson surface reconstruction~\cite{kazhdan-2006-poissonsurfacereconstruction} stage.

Poisson surface reconstruction first recovers an underlying occupancy field $\chi:\IR^3\mapsto [0,1]$
and applies a marching algorithm on $\chi$, which allows for a much better mesh reconstruction than the density function. This approach allows for high scalability as the marching algorithm is applied only in voxels located close to the point cloud. 

To estimate $\chi$, Poisson surface reconstruction discretizes the scene into $2^D \times 2^D \times 2^D$ cells by adapting an octree with depth $D\in \IN$ to the input samples.  $D$ is a hyperparameter provided by the user: The higher $D$, the higher the resolution of the mesh.

By default, SuGaR~\cite{guedon2023sugar} uses a large depth $D=10$ for any scene, as it guarantees a high level of details. However, if the resolution is too high with respect to the complexity of the geometry and the size of the details in the scene, the shapes of the Gaussians become visible as ellipsoidal bumps on the surface of the mesh, and create incorrect bumps or self-intersections. More importantly, holes can also appear in the geometry when $D$ is too large with regards to the density of the Gaussians and the sampled point cloud.


We therefore introduce a method to automatically select $D$. A simple strategy would be to adjust the depth of the octree such that the size of a cell is approximately equal or larger than the average size of the Gaussians in the scene, normalized by the spatial extent of the point cloud used for reconstruction. Unfortunately, this does not work well in practice: We found that whatever the scene (real or synthetic) or the number of Gaussians to represent it, Gaussian Splatting optimization systematically converges toward a varied collection of Gaussian sizes, so that there is no noticeable difference or pattern in the distribution of sizes between scenes.

We noticed that the distance between Gaussians is much more representative of the geometrical complexity of the scene and thus a reliable cue to fix $D$. Indeed, a large but very detailed shape can be reconstructed using Gaussians with large size, if these Gaussians are close to each other. On the contrary, whatever their size, if the centers of the Gaussian are too far from each other, then the rendered geometry will look rough. 

Consequently, to first evaluate the geometrical complexity of a scene, we propose to compute, for each Gaussian $g$ in the scene, the distance between $g$ and its nearest neighbor Gaussian. We use these distances to define the following geometrical complexity score $CS$:
\begin{equation}
    CS = Q_{0.1} \left( \left\{\min_{g'\neq g} \frac{\|\mu_g - \mu_{g'}\|_2}{L}\right\}_{g\in \calG} \right) \> ,
    \label{eq:complexity_score}
\end{equation}
where $\calG$ is the set of all 3D Gaussians in the scene, $L$ is the length of the longest edge of the bounding box of the point cloud to use in Poisson reconstruction, and $Q_{0.1}$ is the function that returns the 0.1-quantile of a list. We use the 0.1-quantile rather than the average because Gaussians that have a neighbor close to them generally encode details in the scene, which provide a much more reliable and less noisy criterion than using the overall average. We also use a quantile rather than a minimum to be robust to extreme values. In short, this complexity score $CS$ is a canonical distance between the closest Gaussians in the scene, i.e., the distance between neighbor Gaussians that reconstruct details in the scene.

Since the normalized length of a cell in the octree is $2^{-\bar{D}}$ and this score represents a canonical normalized distance between Gaussians representing details in the scene, we can compute a natural optimal depth $\bar{D}$ for the Poisson reconstruction algorithm:
\begin{equation}
    \bar{D} = \lfloor -\log_2 \left(\gamma \times CS\right) \rfloor \> ,
    \label{eq:optimal-depth}
\end{equation}
where $\gamma > 0$ is a hyperparameter that does not depend on the scene and its geometrical complexity. This formula guarantees that the size of the cells is as close as possible but greater than $\gamma \times CS$. Decreasing the value of $\gamma$ increases the resolution of the reconstruction. But for a given $\gamma$, whatever the dataset or the complexity of the scene, this formula enforces the scene to be reconstructed with a similar level of smoothness. 

Choosing $\gamma$ is therefore much easier than having to tune $D$ as it is not dependent on the scene. 
In practice, we use $\gamma=100$ for all the scenes. Our experiments validates that this method to fix $D$ results in greater rendering performance.

\section{Initializing the frosting layer}

\subsection{Sampling Gaussians in the frosting layer} 

\subsubsection{Sampling more Gaussians in thicker parts of the frosting.} For a given budget $N$ of Gaussians provided by the user, we initialize $N$ Gaussians in the scene by sampling $N$ 3D centers $\mu_g$ in the frosting layer. Specifically, for sampling a single Gaussian, we first randomly select a prismatic cell with a probability proportional to its volume. Then, we sample random coordinates that sum up to 1. 
This sampling allows for allocating more Gaussians in areas with fuzzy and complex geometry, where more volumetric rendering is needed. However, flat parts  in the layer may also need a large number of Gaussians to recover texture details. Therefore, in practice, we instantiate $N/2$ Gaussians with uniform probabilities in the prismatic cells, and $N/2$ Gaussians with probabilities proportional to the volume of the cell.

\subsubsection{Contracting volumes in unbounded scenes.} In real unbounded scenes, 3D Gaussians located far away from the center of the scene can have a significantly large volume despite their limited participation in the final rendering. This can lead to an unnecessarily large number of Gaussians being sampled in the frosting layer far away from the training camera poses. To address this issue, we propose distributing distant Gaussians proportionally to disparity (inverse distance) rather than distance.

When sampling Gaussians in practice, we start by contracting the volumes of the prismatic cells. We achieve this by applying a continuous transformation $f:\IR^3 \rightarrow \IR^3$ to the vertices of the outer and inner bounds of the frosting layer. 
Then, we compute the volumes of the resulting ``contracted'' prismatic cells and use these adjusted volumes for sampling Gaussians within the frosting layer, as previously described. The transformation function $f$ aims to contract the volume of prismatic cells located far away from the center of the scene. We define $f$ using a formula similar to the contraction transformation introduced in Mip-NeRF~360~\cite{barron2022mipnerf360}:
\begin{equation}
    f(x) = \begin{cases}
        x & \text{if} \>\>\>\> \|x-c\| \leq l\\
        c + l \times \left(2 - \frac{l}{\|x-c\|}\right) \left(\frac{x-c}{\|x-c\|}\right) & \text{if} \>\>\>\> \|x-c\| > l
    \end{cases} \>,
    \label{eq:contraction}
\end{equation}
where $c\in \IR^3$ is the center of the bounding box containing all training camera positions, and $l\in\IR_+$ is equal to half the length of the diagional of the same bounding box. 
We choose the bounding box of the camera positions as our reference scale because both 3D Gaussian Splatting~\cite{kerbl3Dgaussians} and SuGaR~\cite{guedon2023sugar} use this same reference for scaling learning rates and distinguishing foreground from background in unbounded scenes.

\subsection{Avoiding self-intersections in the frosting layer}

In the main paper, we define the inner and outer bounds of the frosting layer by adding inner and outer shifts $\innershift_i$ and $\outershift_i$ to the vertices $\vec{v}_i$ of the base mesh. This results in two bounding surfaces with vertices $\vec{v}_i + \innershift_i$ and $\vec{v}_i + \outershift_i$. 
In practice, we wish to minimize self-intersections within the frosting layer, specifically avoiding prismatic cells intersecting with each other.

While self-intersections do not directly impact rendering quality, they can lead to artifacts during scene editing or animation. Consider the scenario where different cells intersect. In such cases, moving a specific triangle of the base mesh may not affect all Gaussians intersecting the surrounding cell: Some Gaussians may belong to prismatic cells associated with different triangles, resulting in artifacts due to their failure to follow local motion or scene edits.

To mitigate self-intersections, we adopt an indirect approach for initializing the shifts $\innershift$ and $\outershift$. 
Instead of using the final computed values directly, we start with shifts equal to zero and progressively increase them until reaching their final values. 
As soon as an inner vertex (or outer vertex) of a prismatic cell is detected to intersect another cell, we stop further increases in its inner shift (or outer shift). 
This straightforward process significantly reduces self-intersections in the frosting layer while maintaining rendering performance.

By following this approach, we ensure that the frosting layer remains free from unwanted artifacts while preserving efficient rendering capabilities.

\section{Adjusting Gaussians' parameters for edition}

When editing or animating the scene, we automatically adjust Gaussians' parameters. Specifically, in a given prismatic cell with center $\vec{c}$ and six vertices $\vec{v}_i$ for $0\leq i<6$, we first estimate the local transformation at each vertex $\vec{v}_i$ by computing the rotation and rescaling of the vector $(c - \vec{v}_i)$. 

To compute the local rotations at vertex $\vec{v}_i$, we use an axis-angle representation where the axis angle is the normalized cross-product between the previous and the current values of the vector $(c - \vec{v}_i)$.
The local rescaling transformation at vertex $\vec{v}_i$ is computed as the transformation that scales along axis $(c - \vec{v}_i)$ with the appropriate factor but leaves other axes unchanged.

To update the scaling factors and rotation of a Gaussian $g$, we first apply each of these six transformations on the three main axes of the Gaussian. We then average the resulting axes using the barycentric coordinates of the center $\mu_g$ of the Gaussian. We finally orthonormalize the three resulting axes.

\end{document}